\documentclass{article} % For LaTeX2e
\usepackage{template,times}

\newcommand{\msg}[2]{m_{#1\rightarrow #2}}
\newcommand{\N}{\mathcal{N}}

\usepackage{amsmath, amsthm,amsfonts,bm, amssymb, upgreek, array}
\theoremstyle{remark}

\newtheorem{building_block}{Building Block}

\theoremstyle{definition}

\def\eqref#1{equation~\ref{#1}}
\def\1{\bm{1}}

\def\ra{{\textnormal{a}}}
\def\rb{{\textnormal{b}}}
\def\rc{{\textnormal{c}}}

\def\ro{{\textnormal{o}}}
\def\rp{{\textnormal{p}}}

\def\rs{{\textnormal{s}}}

\def\rv{{\textnormal{v}}}
\def\rw{{\textnormal{w}}}
\def\rx{{\textnormal{x}}}
\def\ry{{\textnormal{y}}}
\def\rz{{\textnormal{z}}}
\def\rZ{{\textnormal{Z}}}

\def\rvtheta{{\boldsymbol{\uptheta}}}
\def\rva{{\mathbf{a}}}
\def\rvb{{\mathbf{b}}}

\def\rvo{{\mathbf{o}}}

\def\rvv{{\mathbf{v}}}

\def\rvx{{\mathbf{x}}}
\def\rvy{{\mathbf{y}}}
\def\rvz{{\mathbf{z}}}

\def\rmW{{\mathbf{W}}}

\def\vmu{{\bm{\mu}}}
\def\vtheta{{\bm{\theta}}}
\def\va{{\bm{a}}}

\def\vc{{\bm{c}}}

\def\vo{{\bm{o}}}

\def\vt{{\bm{t}}}

\def\vv{{\bm{v}}}

\def\vx{{\bm{x}}}
\def\vy{{\bm{y}}}

\def\mW{{\bm{W}}}

\DeclareMathAlphabet{\mathsfit}{\encodingdefault}{\sfdefault}{m}{sl}
\SetMathAlphabet{\mathsfit}{bold}{\encodingdefault}{\sfdefault}{bx}{n}

\def\sG{{\mathbb{G}}}

\newcommand{\E}{\mathbb{E}}

\newcommand{\R}{\mathbb{R}}

\newcommand{\KL}{D_{\mathrm{KL}}}

\usepackage{tikzit}
\usetikzlibrary{fit, positioning}
\usetikzlibrary{decorations.pathreplacing}
\usetikzlibrary{arrows.meta,positioning,calc}
\tikzstyle{factor}=[fill=black, draw=black, shape=rectangle]
\tikzstyle{variable_node}=[fill=white, draw=black, shape=circle]
\tikzstyle{known_variable}=[fill={rgb,255: red,200; green,200; blue,200}, draw=black, shape=circle, font=\small, minimum size=10pt]

\tikzstyle{incoming_message}=[fill=none, ->]
\tikzstyle{custom_arrow}=[->, line width=1.2pt, color=red]
\definecolor{iclr_light_green}{RGB}{211, 223, 58}
\definecolor{iclr_dark_green}{RGB}{93, 188, 122}
\definecolor{iclr_blue}{RGB}{41, 197, 231}
\definecolor{oxygenated_red}{RGB}{190, 32, 47}
\definecolor{deoxygenated_blue}{RGB}{61, 121, 189}
\definecolor{matte_green}{RGB}{57,173,72}

\usepackage{hyperref}
\usepackage{url}
\usepackage{subcaption}
\usepackage{multirow}
\usepackage{nicematrix}
\usepackage{booktabs}
\usepackage{float}
\usepackage{enumitem}
\usepackage{hyperref}
\usepackage{cleveref}
\usepackage{listings}
\usepackage{footmisc}

\title{\centering Approximate Message Passing for Bayesian Neural Networks}

\newcommand*\circled[1]{\tikz[baseline=(char.base)]{
    \node[shape=circle,draw,inner sep=0.5pt] (char) {#1};}}

\iclrfinalcopy % Uncomment for camera-ready version, but NOT for submission.
\begin{document}
\maketitle
\vspace{-1.5cm}
\begin{center}
\textbf{Romeo Sommerfeld\footnote{Equal Contribution\label{equal_contribution}}\footnote{Hasso Plattner Institute, University of Potsdam\label{institution}} \hspace{12pt}Christian Helms}\footref{equal_contribution}\footref{institution} \hspace{12pt}Ralf Herbrich\footref{institution}
\end{center}
\vspace{1cm}
% Limit: 5000 characters. Some random example paper (https://openreview.net/pdf?id=EhrzQwsV4K) only used 1739 characters.
\begin{abstract}

% Structure of the Abstract:
% - Aim / Problem / What the paper does:
%  - Deep learning focuses primarily on accuracy, but BNNs are important for: uncertainty quantification, data efficiency, interpretability, and more.
%  - Despite recent advances, BNNs have struggled to achieve wider adoption in practice, calling for a wider diversity of methods that can be tried.
%  - We develop a new framework based on message passing

% - Methodology / Materials
%  - We evaluate uncertainty on MNIST and synthetic data
%  - Runtime tests

% - Results / Achievements:
%  - 1. Excellent low-data performance. LeNet-5 achieves 94.62% accuracy on n=640, SGD only 22.15%
%  - 2. Uncertainty metrics clearly outperform SGD, and we find generally accurate ood-certainties on synthetic data

% - Contribution / Implication / Application
%  - First scalable MP framework that doesn't double-count (EBP, PBP, Lucibello all double-count)
%  - First time to apply MP to CNNs
%  -> We push the boundary of MP-based BNNs

% - Limitations and Future Work
%  - Scale and computational speed are still worse than VI. The largest MLP we trained had only 5.6mio parameters. Training is much slower than SGD, but inference is only 1.6x slower
%  - Larger datasets, expanded architectures, real-world tasks

% Do we want to specifically attack VI ("existing Variational Inference (VI) methods often struggle...")? I feel like that's a bit too much
Bayesian neural networks (BNNs) offer the potential for reliable uncertainty quantification and interpretability, which are critical for trustworthy AI in high-stakes domains. However, existing methods often struggle with issues such as overconfidence, hyperparameter sensitivity, and posterior collapse, leaving room for alternative approaches. In this work, we advance message passing (MP) for BNNs and present a novel framework that models the predictive posterior as a factor graph. To the best of our knowledge, our framework is the first MP method that handles convolutional neural networks and avoids double-counting training data, a limitation of previous MP methods that causes overconfidence. We evaluate our approach on CIFAR-10 with a convolutional neural network of roughly 890k parameters and find that it can compete with the SOTA baselines AdamW and IVON, even having an edge in terms of calibration. On synthetic data, we validate the uncertainty estimates and observe a strong correlation (0.9) between posterior credible intervals and its probability of covering the true data-generating function outside the training range. While our method scales to an MLP with 5.6 million parameters, further improvements are necessary to match the scale and performance of state-of-the-art variational inference methods.

\end{abstract}

\section{Introduction}

Deep learning models have achieved impressive results across various domains, including natural language processing \citep{vaswani2023attentionneed}, computer vision \citep{ravi2024sam2segmentimages}, and autonomous systems \citep{bojarski2016endendlearningselfdriving}. Yet, they often produce overconfident but incorrect predictions, particularly in ambiguous or out-of-distribution scenarios. Without the ability to effectively quantify uncertainty, this can foster both overreliance and underreliance on models, as users stop trusting their outputs entirely \citep{zhang2024beyond}, and in high-stakes domains like healthcare or autonomous driving, its application can be dangerous \citep{henne2020benchmarking}. To ensure safer deployment in these settings, models must not only predict outcomes but also express how uncertain they are about those predictions to allow for informed decision-making.

Bayesian neural networks (BNNs) offer a principled way to quantify uncertainty by capturing a posterior distribution over the model's weights, rather than relying on point estimates as in traditional neural networks. This allows BNNs to express epistemic uncertainty, the model's lack of knowledge about the underlying data distribution. Current methods for posterior approximation largely fall into two categories: sampling-based methods, such as Hamiltonian Monte Carlo (HMC), and deterministic approaches like variational inference (VI). While sampling methods are usually computationally expensive, VI has become increasingly scalable \citep{ivon_shen2024variationallearningeffectivelarge}. However, VI is not without limitations: It often struggles with overconfidence \citep{papamarkou2024positionbayesiandeeplearning}, and it can struggle to break symmetry when multiple modes are close \citep{zhang2018advancesvariationalinference}. Mean-field approaches, commonly used in VI, are prone to posterior collapse \citep{kurle2022detrimentaleffectinvarianceslikelihood,coker2022widemeanfieldbayesianneural}. Additionally, VI often requires complex hyperparameter tuning \citep{vogn_osawa2019practicaldeeplearningbayesian}, which complicates its practical deployment in real-world settings. These challenges motivate the need for alternative approaches that can potentially address some of the shortcomings of VI while maintaining its scalability.

In contrast, message passing (MP) \citep{minka_ep} is a probabilistic inference technique that suffers less from these problems. 
Belief propagation \citep{sumproduct_kschischang}, the basis for many MP algorithms, integrates over variables of a joint density $p(\rx_1, \ldots, \rx_n)$ that factorize into a product of functions $f_j$ on subsets of random variables $\rx_1, \ldots, \rx_n$. 
The corresponding factor graph is bipartite and connects these factors $f_j$ with the variables they depend on. 
The following recursive equations yield a computationally efficient algorithm to compute all marginals $p(\rx_i)$ for acyclic factor graphs:
\begin{align*}
    p(\rx) &= \prod_{f \in N_\rx} m_{f\rightarrow \rx} (\rx) & m_{f\rightarrow \rx}(\rx) &= \int f(N_f)\prod_{\ry \in N_f\setminus \{\rx\}} \msg{\ry}{f}(\ry)\,d(N_f \setminus \{\rx\})
\end{align*}
where $N_v$ denotes the neighborhood of vertex $v$ and $m_{\ry \rightarrow f}(\ry) = \prod_{f^\prime \in N_\ry \setminus \{f\}} m_{f^\prime\rightarrow \ry} (\ry)$. Since exact messages are often intractable and factor graphs are rarely acyclic, belief propagation typically cannot be applied directly. Instead, messages $m_{f\rightarrow X}(\cdot)$ and marginals $p_X(\cdot)$ are typically approximated by some family of distributions that has few parameters (e.g., Gaussians). However, applying message passing (MP) in practice presents two main challenges for practitioners: the need to derive (approximate) message equations when $m_{f\rightarrow \rx}$ falls outside the approximating family, and the complexity of implementing MP compared to other methods.

% One more paragraph that makes a good case for why message passing is a promising and underexplored method for BNNs. Mention that we are advancing the SOTA of MP
% \begin{enumerate}
%     \item Intro to MP and Factor Graphs + their background (briefly)
%     \item Why does MP have the potential to make great BNNs?
%     \item Current approaches (briefly) ($\rightarrow$ EBP, PBP, Lucibello)
%     \item Challenges of MP (which creates a good bridge towards our list of contributions)
%     \begin{enumerate}
%         \item Lack of scalable, performant implementations ($\rightarrow$ We present our Julia implementation)
%         \item Need to derive message equations ($\rightarrow$ Message Equations in Appendix)
%         \item More challenges? Why hasn't it happened before?
%     \end{enumerate}
% \end{enumerate}
% (Modularity to model some covariances and ignore others, well-suited for distributed settings or tasks with a lot of structure, ...?)

% 1. Propose scalable framework for BNNs with message passing that avoids double-counting. We derive lots of message equations for a variety of factors that will be useful for MP beyond BNNs.
% 2. Implementation in Julia on CPU/GPU (which is significantly different from the theoretical model) and demonstrate that it scales to CNNs and large MLPs.
% 3. Compare metrics (uncertainty, accuracy, performance) on MNIST and use synthetic data to analyze quality of uncertainty approximation.

We summarize our contributions as follows:
\begin{enumerate}[leftmargin=*,itemsep=0.01mm,topsep=-0.5\parskip]
    \item We propose a scalable message-passing framework for Bayesian neural networks and derive message equations for various factors, which can benefit factor graph modeling across domains.
    \item We implement our method in Julia for both CPU and GPU, and demonstrate its scalability to convolutional neural networks (CNNs) and large multilayer perceptrons (MLPs).
    \item We evaluate on CIFAR-10 and find that our method is competitive with the SOTA baselines AdamW and IVON, even having an edge in terms of calibration while requiring no hyperparameter tuning.  
\end{enumerate}
To the best of our knowledge, this is the first MP method to handle CNNs and to avoid double-counting training data, thereby preventing overconfidence and, eventually, posterior collapse. While our methods scales to an MLP with 5.6 million parameters, further refinements are necessary to match the scale and performance of state-of-the-art VI methods.

\subsection{Related Work}
As the exact posterior is intractable for most practical neural networks, approximate methods are essential for scalable BNNs. These methods generally fall into two categories: sampling-based approaches and those that approximate the posterior with parameterized distributions.

% Some relevant papers:
% - https://arxiv.org/pdf/2202.11670 (see chatgpt text below)
% - Papamarkou \cite{papamarkou2023approximateblockedgibbssampling}: Only 8180 parameters for an MNIST MLP!

% General propblems:
% - Slow performance
% - When to stop (how is the mixing going?)
% - Afterwards you have an array of network weights, which costs memory and performance during prediction

% - Gradient extensions, such as SGLD offer approximate sampling (instead of exact, such as Metropolis Hastings) at much faster speed
% - Bayesian Dark Knowledge solves the inference cost problem, but MCMC methods (including SGLD)
% - However, SGLD is also still too expensive / ineffective for large neural networks (see https://arxiv.org/pdf/2107.04562 by Khan)
\textbf{Markov Chain Monte Carlo}
(MCMC) methods attempt to draw representative samples from posterior distributions. Although methods such as Hamiltonian Monte Carlo are asymptotically exact, they become computationally prohibitive for large neural networks due to their high-dimensional parameter spaces and complex energy landscapes \citep{coker2022widemeanfieldbayesianneural}. An adaptation of Gibbs sampling has been scaled to MNIST, but on a very small network with only 8,180 parameters \citep{papamarkou2023approximateblockedgibbssampling}. Approximate sampling methods can be faster but still require a large number of samples, which complicates both training and inference. Although approaches like knowledge distillation \citep{korattikara2015bayesiandarkknowledge} attempt to speed up inference, MCMC remains generally too inefficient for large-scale deep learning applications \citep{khan2024bayesianlearningrule}.

\textbf{Variational Inference} 
(VI) aims to approximate the intractable posterior distribution $p(\theta\,|\,\mathcal{D})$ by a variational posterior $q(\theta)$. The parameters of $q$ are optimized using gradients with respect to an objective function, which is typically a generalized form of the reverse KL divergence $\KL\left[\,{q(\theta) \,\|\, p(\theta\,|\,\mathcal{D})}\,\right]$. Early methods like \citep{graves2011_vi} and Bayes By Backprop \citep{blundell2015weightuncertaintyneuralnetworks} laid the foundation for applying VI to neural networks, but suffer from slow convergence and severe underfitting, especially for large models or small dataset sizes \citep{vogn_osawa2019practicaldeeplearningbayesian}. More recently, VOGN \citep{vogn_osawa2019practicaldeeplearningbayesian} achieved Adam-like results on ImageNet LSVRC by applying a Gauss-Newton approximation to the Hessian matrix. IVON \citep{ivon_shen2024variationallearningeffectivelarge} improved upon VOGN by using cheaper Hessian approximations and training techniques like gradient clipping, achieving Adam-like performance on large-scale models such as GPT-2 while maintaining similar runtime costs. Despite recent advancements, VI continues to face challenges such as overconfidence, posterior collapse, and complex hyperparameter tuning (see introduction), motivating the exploration of alternative approaches \citep{zhang2018advancesvariationalinference}.

% - EBP is expectation propagation -> reverse KL divergence (similar to VI)
% - PBP uses moment matching (using gradients) -> normal KL divergence!
% - Lucibello is unclear, but I would assume it uses normal KL divergence. Badly written paper...
\textbf{Message Passing for Neural Networks}: Message passing is a general framework that unifies several algorithms \citep{sumproduct_kschischang, minka_ep}, but its direct application to neural networks has been limited. Expectation backpropagation (EBP) \citep{expectation_backpropagation} approximates the posterior of 3-layer MLPs by combining expectation propagation, an approximate message passing algorithm, with gradient backpropagation. Similarly, probabilistic backpropagation (PBP) \citep{hernándezlobato2015probabilisticbackpropagationscalablelearning} combines belief propagation with gradient backpropagation and was found to produce better approximations than EBP \citep{Ghosh_Delle_Fave_Yedidia_2016}. However, PBP treats the data as new examples in each consecutive epoch (double-counting), which makes it prone to overconfidence. Furthermore, EBP and PBP were both only deployed on small datasets and rely on gradients instead of pure message passing. In contrast, \citet{Lucibello_2022} applied message passing to larger architectures by modeling the posterior over neural network weights as a factor graph, but faced posterior collapse to a point measure due to also double-counting data. Their experiments were mostly restricted to three-layer MLPs without biases and with binary weights. Our approach builds on this by introducing a message-passing framework for BNNs that avoids double-counting, scales to CNNs, and effectively supports continuous weights.

\section{Theoretical Model}
\begin{figure}[hbt!]
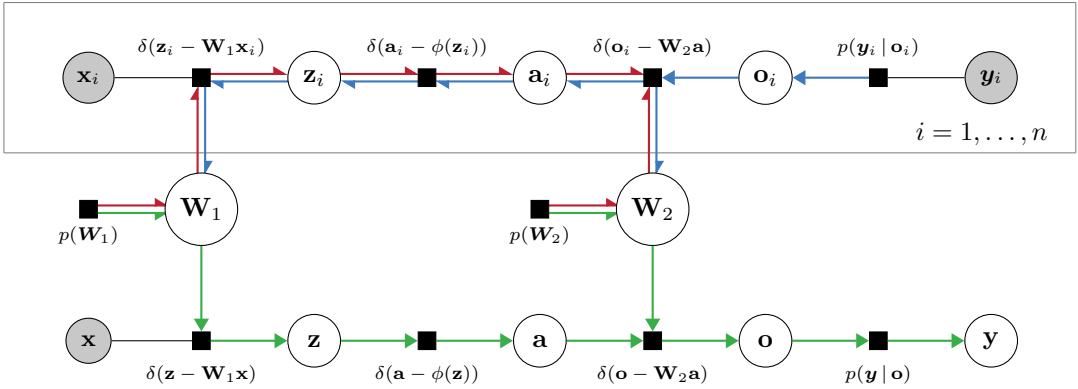

    \centering
    \ctikzfig{fg_iteration_order}
    \caption{Conceptual vector-valued factor graph for a simple MLP. Each training example has its own "branch" (a copy of the network), while predictions for an unlabeled input $\vx$ are computed on a separate prediction branch. All branches are connected by the shared model parameters. Grayed-out variables are conditioned on (observed). Colored arrows indicate the three iteration orders: a \textcolor{oxygenated_red}{forward} / \textcolor{deoxygenated_blue}{backward} pass on training examples,
    % (like (de)oxygenated blood flowing in(out))
    and a forward pass for \textcolor{matte_green}{prediction}. }
    \label{fig:fg_iteration_order}
\end{figure}

Our goal is to model the predictive posterior of a BNN as a factor graph and find a Gaussian approximation of the predictive posterior via belief propagation.     
Essentially, factor graphs are probabilistic modelling tools for approximating the marginals of joint distributions, provided that they factorize sufficiently.
For a more comprehensive introduction on factor graphs and the sum-product algorithm, refer to \cite{sumproduct_kschischang} 
BNNs, on the other hand, treat the parameters $\rvtheta$ of a model $f_{\rvtheta}:\R^d\longrightarrow \R^o$ as random variables with prior beliefs $p(\rvtheta)$.
Given a training dataset $\mathcal{D} = \{\vx_i,\vy_i\}_{i=1}^n$ of i.i.d. samples, a likelihood relationship $p(\vy\,|\,\vx,\rvtheta) = \,p(\vy\,|\,f_{\rvtheta}(\vx))$, and a new input sample $\vx$, the goal is to approximate the predictive posterior distribution $p(\vy\,|\,\vx,\mathcal{D})$, which can be written as:
\begin{equation}
    \label{equation:predictive_posterior}
    p(\vy\,|\,\vx,\mathcal{D}) = \int p(\vy\,|\,\vx,\rvtheta)\,p(\rvtheta\,|\,\mathcal{D})\,d\rvtheta. 
\end{equation}
This means that the density of the predictive posterior is the expected likelihood under the posterior distribution $p(\rvtheta\,|\,\mathcal{D})$, which is proportional\footnote{with a proportionality constant of $1/p(\mathcal{D})$} to the product of the prior and dataset likelihood:
\begin{equation}
    p(\rvtheta\,|\,\mathcal{D}) \propto p(\rvtheta) \prod_{i=1}^n p(\vy_i\,|\,f_\rvtheta(\vx)).
\end{equation}
The integrand in \Cref{equation:predictive_posterior} exhibits a factorized structure that is well-suited to factor graph modeling.
However, directly modelling the relationship $\vo = f_\rvtheta(\vx)$ with a single Dirac delta factor $\delta(\vo - f_\rvtheta(\vx))$ does not yield feasible message equations.
Therefore we model the neural network at scalar level by introducing intermediate latent variables connected by elementary Dirac delta factors.
\Cref{fig:fg_iteration_order} illustrates this construction for a simple MLP with independent weight matrices a priori.
While the abstract factor graph in the figure uses vector variables for simplicity, we actually derive message equations where each vector component is treated as a separate scalar variable, and all Dirac deltas only depend on scalar variables.
For instance, if $d=2$, the conceptual factor $\delta(\rvo - \rmW_2\rva)$ is replaced by four scalar factors: $\delta(\rp_{jk} - \rw_{jk}\ra_{k})$ for $j,k=1,2$, with intermediate variables $\rp_{jk}$, and two factors $\delta(\ro_j - (\rp_{j1} + \rp_{j2}))$.  
By multiplying all factors in this expanded factor graph and integrating over intermediate results, we obtain a function in $\vx,\vy,\rvtheta$ that is proportional to the integrand in \Cref{equation:predictive_posterior}.
Hence, the marginal of the unobserved target $\rvy$ is proportional to $p(\vy\,|\,\vx,\mathcal{D})$.
When $\rvy$ connects to only one factor, its marginal matches its incoming message.

\section{Approximations}
Calculating a precise representation of the message to the target of an unseen input is intractable for large networks and datasets.
The three primary reasons are, that a) nonlinearities and multiplication produce highly complex exact messages which are difficult to represent and propagate, b) the enormous size of the factor graph for large datasets, and c) the presence of various cycles in the graph.
These challenges shape the message approximations as well as the design of our training and prediction procedures, which we address in the following sections.

% To address a), we approximate messages with a parameterized class of functions, desiring closure under pointwise multiplication. We choose positive scalar multiples of one-dimensional Gaussian densities, as they fulfill this property. From the identity $\exp(x)\exp(y) = \exp(x + y)$, and noting that for $s_1, s_2 > 0$ and $\mu_1, \mu_2 \in \R$, the function $s_1(x-\mu_1)^2 + s_2(x-\mu_2)^2$ can be written as $s(x - \mu)^2 + c$, we use a natural re-parameterization. 

% Given a Gaussian $\N(\mu, \sigma^2)$, we define the precision $\rho = 1/\sigma^2$ and precision-mean $\tau = \mu/\sigma^2$. Collectively, $(\tau, \rho)$ are the natural parameters, and $\sG(x; \tau, \rho) := \N(x; \mu, \sigma^2)$. For $\mu_1, \mu_2 \in \mathbb{R}$ and $\sigma_1, \sigma_2 > 0$, with corresponding natural parameters $\rho_i = 1/\sigma_i^2$ and $\tau_i = \mu_i \rho_i$, multiplying Gaussian densities simplifies as:

\subsection{Approximating Messages via Gaussian Densities}
To work around  the highly complex exact messages, we approximate them with a parameterized class of functions. We desire this class to be closed under pointwise multiplication, as variable-to-factor messages are the product of incoming messages from other factors.
We choose positive scalar multiples of one-dimensional Gaussian densities as our approximating family.
Their closedness follows immediately from the exponential function's characteristic identity $\exp(x)\exp(y) = \exp(x + y)$ and the observation that for any $s_1,s_2>0$ and $\mu_1,\mu_2\in\R$, the function $s_1(x-\mu_1)^2 + s_2(x-\mu_2)^2$ in $x$ can be represented as $s(x - \mu)^2 + c$ for some $s > 0$ and $\mu_,c\in\R$. 
The precise relation between two scaled Gaussian densities and its product can be neatly expressed with the help of the so-called natural (re-)parameterization. 
Given a Gaussian $\N(\mu, \sigma^2)$, we call $\rho = 1/\sigma^2$ the precision and $\tau = \mu / \sigma^2$ the precision-mean.
Collectively, $(\tau,\rho)$ are the Gaussian's natural parameters, $\sG(x; \tau, \rho) := \N(x; \mu, \sigma^2),~x\in\R.$
For $\mu_1,\mu_2\in\mathbb{R}$ and $\sigma_1,\sigma_2 > 0$ with corresponding natural parameters $\rho_i = 1/\sigma_i^2$ and $\tau_i = \mu_i \rho_i,\,i=1,2$, multiplying Gaussian densities simplifies to:
\begin{equation}
    \label{equation:multiplying_natural_parameter_gaussians}
    \sG(x; \tau_1, \rho_1) \cdot \sG(x; \tau_2, \rho_2) = \N(\mu_1; \mu_2, \sigma_1^2 + \sigma_2^2) \cdot\sG(x; \tau_1 + \tau_2, \rho_1 + \rho_2)
\end{equation}
for all $x\in\R$.
In other words, multiplying Gaussian densities simplifies to the pointwise addition of their natural parameters, aside from a multiplicative constant. Since we are only interested in the marginals, which are re-normalized, this constant does not affect the final result. Therefore, we can safely ignore these multiplicative constants and only keep track of the Gaussian's parameters.
% That means, multiplication of Gaussian densities reduces to pointwise addition of their natural parameters up to a multiplicative constant.
% But for our purposes we can ignore this constant as it does not affect other messages, except for this very multiplicative constant, due to the linearity of the integral.
% At the end of the day, we are only interested in the approximation of marginals by normalized Gaussian densities parameterized with natural parameters; and multiplicative constants do not influence this representation.   

Now we present our message approximations for three factor types, each representing a deterministic relationship between variables: \circled{1} the sum of variables weighted by constants, \circled{2} the application of a nonlinearity, and \circled{3} the multiplication of two variables.
As we model the factor graph on a scalar level, these three factors suffice to model complex modern network architectures such as ConvNeXt \cite{ConvNeXt}\footnote{with the exception of layer normalization, which can be substituted by orthogonal initialization schemes \cite{DynamicalIsometryAndMeanFieldTheoryOfCNNs} or specific hyperparameters of a corresponding normalized network \cite{UN-NormalizedConvolution}}.
In \ref{appendix:message_equation_table}, we provide a comprehensive table of message equations, including additional factors for modeling training labels.

\textbf{Weighted Sum}: The density transformation property of the Dirac delta allows us to compute the exact message without approximation.
For the relationship $\rs = \vc^\intercal \rvv$ modeled by the factor $f := \delta(s - \vc^\intercal \vv)$, the message 
$$\msg{f}{\rs}(s) = \int \delta(s - \vc^\intercal \vv)\prod_{i=1}^k\msg{\rv_i}{f}(v_i)\,dv_1\dots v_k$$
is simply the density of $\vc^\intercal \rvv$, where $\rvv \sim \prod_{i=1}^k\msg{\rv_i}{f}(v_i).$  
If $\msg{\rv_i}{f}(v_i) = \N(v_i; \mu_i, \sigma^2_i)$ are Gaussian, then $\rvv \sim \N(\vmu, \text{diag}(\boldsymbol{\sigma}^2))$ and $\msg{f}{s}(s)$ becomes a scaled multivariate Gaussian:
$$\msg{f}{\rs}(s) = \N(s; \vc^\intercal \vmu, (\vc^2)^\intercal \boldsymbol{\sigma}^2).$$
The backward messages $\msg{f}{\rv_i}$ can be derived similarly without approximation.

\textbf{Nonlinearity}: We model the application of a nonlinearity $\phi:\R \rightarrow \R$ as a factor $f := \delta(\ra - \phi(\rz))$. However, the forward and backward messages are problematic and require approximation–even for well-behaved, injective $\phi$ such as $\text{LeakyReLU}_\alpha$:
\begin{gather*}
    \msg{\ra}{f}(\ra) = \text{pdf}_{\phi(\rZ)}(\ra) ~~ \text{for } \rZ \sim \N \\
    \msg{f}{\rz}(\rz) = \int \delta(\ra - \phi(\rz))\cdot \msg{\ra}{f}(\ra)\,d\ra = \msg{\ra}{f}(\phi(z)) = \N(\phi(\rz);\mu_\ra,\sigma_\ra^2).
\end{gather*}
For values of $\alpha \neq 1$, the forward message is non-Gaussian and the backward message does not even integrate to 1. For $\text{ReLU}$ ($\alpha=0)$, it is clearly not even integrable. Instead, we use \emph{moment matching} to fit a Gaussian approximation.
Given any factor $f$ and variable $\rv$, we can approximate the message $\msg{f}{\rv}$ with a Gaussian if the moments $m_k := \int \rv^k\msg{f}{\rv}(\rv)\,d\rv$ exist for $k=0,1,2$ and can be computed efficiently:
\begin{gather*}
    \msg{f}{\rv}(\rv) = \N(\rv; m_1 / m_0, m_2/m_0 - (m_1 / m_0)^2) \tag{Direct approximation}
\end{gather*}
However, direct moment matching of the message is impossible for non-integrable messages or when the $m_k$ are expensive to find. Instead, we can apply moment matching to the updated \emph{marginal} of $\rv$. Let $m_0$, $m_1$, $m_2$ be the moments of the "true" marginal
$$ m(\rv) = \int f(\rv, \rv_1, ..., \rv_k) ~~ d\rv_1 ... d\rv_k ~\cdot~ \prod_i \msg{g_i}{\rv}(\rv), $$
which is the product of the true message from $f$ and the approximated messages from other factors $g_i$. Then we can approximate $m$ with a Gaussian and obtain a message approximation
\begin{gather*}
    \label{equation:marginal_approximation}
    \tag{Marginal approximation}
    \msg{f}{\rv}(\rv) := \N(\rv;\mu_\rv, \sigma^2_\rv) / \msg{\rv}{f}(\rv)
\end{gather*}
which approximates $\msg{f}{\rv}$ so that it changes $\rv$'s marginal in the same way as the actual message.\footnote{This is the central idea behind expectation propagation as defined in \cite{minka_ep}.}
Since $\msg{\rv}{f}(\rv)$ is a Gaussian density, we can compute $\msg{f}{\rv}(\rv)$ efficiently by applying Gaussian division in natural parameters, similar to \Cref{equation:multiplying_natural_parameter_gaussians}.

For $\text{LeakyReLU}_\alpha$, we found efficient direct and marginal approximations that are each applicable to both the forward and backward message when $\alpha \neq 0$. The marginal approximation remains applicable even for the ReLU case of $\alpha = 0$. We provide detailed derivations in \Cref{appendix:LeakyReLU_derivations}.

\textbf{Product} For the relationship $\rc = \ra \rb$, we employ variational message passing as in \citet{stern2009matchbox}, in order to break the vast number of symmetries in the true posterior of a Bayesian neural network. By combining the variational message equations for scalar products with the weighted sum, we can also construct efficient higher-order multiplication factors such as inner vector products. Refer to \ref{appendix:message_equation_table} for detailed equations.

\subsection{Training Procedure \& Prediction}

In pure belief propagation, the product of incoming messages for any variable equals its marginal under the true posterior. With our aforementioned approximations, we can reasonably expect to converge on a diagonal Gaussian $\check{q}$ that approximates one of the various permutation modes of the true posterior by aligning the first two moments of the marginal.
This concept can be elegantly interpreted through the lens of relative entropy. 
As shown in \ref{appendix:minimization_proof}, among diagonal Gaussians $q(\theta) = q_1(\theta_1) \cdots q_k(\theta_k)$, the relative entropy from (a mode of) the true posterior to $q$ is minimized for $\check{q}$:
\begin{equation}
\label{equation:posterior_approx_minimization}
\check{q} = \underset{q}{\text{argmin}}\,\KL\left[\,{p(\theta\,|\,\mathcal{D})\,|\,q(\theta)}\,\right].
\end{equation}

Another challenge in finding $\check{q}$ arises from cyclic dependencies. In acyclic factor graphs, each message depends only on previous messages from its subtree, allowing for exact propagation. However, our factor graph contains several cycles due to two primary reasons: \circled{1} multiple training branches interacting with shared parameters across linear layers, and \circled{2} the scalar-level modeling of matrix-vector multiplication in architectures with more than one hidden layer. These loops create circular dependencies among messages. To address these challenges, we adopt loopy belief propagation, where belief propagation is performed iteratively until messages converge. While exact propagation works in acyclic graphs, convergence is then only guaranteed under certain conditions (e.g., Simon’s condition \citep{Ihler}) that are difficult to verify. Instead, we pass messages in an iteration order that largely avoids loops by alternating forward and backward passes similarly to deterministic neural networks. Our message schedule is visualized in \Cref{fig:fg_iteration_order}.
% Why do loop-avoiding message equations help? My guess: Because then messages remain "valid" for longer. Assume some message m_m has dependencies on messages m_1, m_2. If an update strongly changes m_1, then our current estimate of m_m diverges strongly from the true message. Updating m_1 thus invalidates m_m. By only iterating long loops, our messages stay valid for longer

\begin{figure}[ht!]
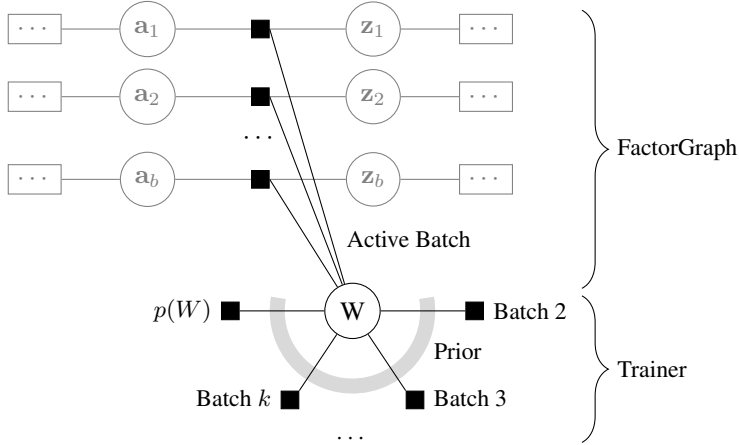

    \centering
    \ctikzfig{batching}
    \caption{A full FactorGraph models all messages for one batch of training examples. To iterate the FactorGraph, we only need one joint message summarizing the prior and all other examples. When switching to a new batch, we aggregate messages from the previous batch and store them in the Trainer.}
    \label{fig:batching}
\end{figure}

\textbf{Batching} As the forward and backward messages depend on each other, we must store them to compute message updates during message passing. Updating our messages in a sweeping "pass" over a branch and running backward passes immediately after the forward pass on the same branch, allows us to store many messages only temporarily, reducing memory requirements. This schedule also ensures efficient propagation of updated messages despite the presence of loops. However, some messages must still be retained permanently\footnote{For example, the backward message of the linear layer is needed to compute the marginal of the inputs, which the forward message depends on.}, leading to significant memory demand when storing them for all $n$ training examples. To address this, we adopt a batching strategy: Instead of maintaining $n$ training branches simultaneously, we update the factor graph using a batch (subset) of $b$ examples at a time. The factor graph then models $b$ messages to the weights $W$, while the messages to $W$ from the remaining (inactive) examples are aggregated into batch-wise products and stored in a trainer object. \Cref{fig:batching} illustrates this setup. When switching batches, we divide the marginals by the batch's old aggregate message and multiply the updated messages into the marginal, ensuring that data is not double-counted.
Within each batch, we iterate through the examples and perform a forward and backward pass on each in sequence. After all examples have been processed once, we call it an "iteration". Depending on the training stage, we either repeat this process within the same batch or move to the next batch. As training progresses, we gradually increase the number of iterations per batch to allow for finer updates as the overall posterior comes closer to convergence.

\textbf{Prediction:}
Ultimately, our goal is to compute the marginal of the unobserved target $\mathbf{y}$ for some unseen input $\mathbf{x}$. Since the prediction branch in \Cref{fig:fg_iteration_order} introduces additional loops, obtaining an accurate approximation would require iterating over the entire factor graph, including the training branches. In neural network terms, this translates to retraining the whole network for every test input. Instead, we pass messages only on the training branches in the batch-wise setup described above. At test time, messages from the training branches are propagated to the prediction branch, but not vice versa. Specifically, messages from the weights to the prediction branch are computed as the product of the prior and the incoming messages from the training branches. This can be interpreted as approximating the posterior over weights, $p(\theta\,|\,\mathcal{D})$, with a diagonal Gaussian $\check{q}(\theta)$ and using it as the prior during inference.

\section{Making It Scale}
% If someone wanted to reimplement our approach, what implementation details should they keep in mind?
% - Performance: Which parts to implement and which parts not to
% - Numerical Stability: ... (See below)
% - Initialization

% \subsection{Initializations}
% - Non-symmetric prior initialization
% - Spectral Scaling and exploding prior variances
% - Orthogonal initialization?

% \subsection{Numerical Stability}
% - Message equations in natural parameters for backward messages
% - Marginal Drift Correction
% - LeakyReLU Guardrails

% - Randomized training order
% - Message Damping

% \subsection{Hardware Acceleration}
% - Implemented using Tullio -> GPU code comes for free

In scaling our approach to deep networks, we encountered several challenges related to computational performance, numerical stability, and weight initialization. The following subsections detail remedies to these problems.

\subsection{Factor Graph Implementation}
% Instead of "individual elements", we could write "individual variables and factors"
While batching effectively reduces memory requirements for large datasets, a direct implementation of a factor graph still scales poorly for deep networks. Explicitly modeling each scalar variable and factor as an instance is computationally expensive. To address this, we propose the following design optimizations: \circled{1} Rather than modeling individual elements of the factor graph, we represent entire layers of the network. Message passing between layers is orchestrated by an outer training loop. \circled{2} Each layer instance operates across all training branches within the active batch, removing the need to duplicate layers for each example. \circled{3} Factors are stateless functions, not objects. Each layer is responsible for computing its forward and backward messages by calling the required functions. 
In this design, layer instances maintain their own state, but message passing and batching are managed in the outer loop. The stateless message equations are optimized for both performance and numerical stability. As a result, the number of layer instances scales linearly with network depth but remains constant regardless of layer size or batch size. This approach significantly reduces computational and memory overhead—our implementation is approximately 300x faster than a direct factor graph model in our tests. Additionally, we optimized our implementation for GPU execution by leveraging Julia’s \texttt{CUDA.jl} and \texttt{Tullio.jl} libraries. Since much of the runtime is spent on linear algebra operations (within linear or convolutional layers), we built a reusable, GPU-compatible library for Gaussian multiplication. This design makes the implementation both scalable and extendable.

\subsection{Numerical Stability}
\label{subsection:numerical_stability}
Maintaining numerical stability in the message-passing process is critical, particularly as model size increases. Backward messages often exhibit near-infinite variances when individual weights have minimal impact on the likelihood. Therefore, we compute them directly in natural parameters, which also simplifies the equations. Special care is needed for LeakyReLU, as its messages can easily diverge. To mitigate this, we introduced guardrails: when normalization constants become too small, precision turns negative, or variance in forward messages increases, we revert to either $\sG(0,0)$ or use moment matching on messages instead of marginals (see \ref{appendix:message_equation_table} for details). Another trick is to periodically recompute the weight marginals from scratch to maintain accuracy. By leveraging the properties of Gaussians, we save memory by recomputing variable-to-factor messages as needed\footnote{Each layer stores factor-to-weight-variable messages and the marginal, which is an aggregate that is continuously updated as individual messages change. To compute a variable-to-factor message, divide the marginal by the factor-to-variable message.}. However, incremental updates to marginals can accumulates errors, so we perform a full recomputation once per batch iteration. Lastly, we apply light message damping through an exponential moving average to stabilize the training, but, importantly, only on the aggregated batch messages, not on the individual messages of the active batch.

\subsection{Weight Priors}
\label{sec:priors}
A zero-centered diagonal Gaussian prior with variance $\sigma^2_p$ is a natural choice for the prior over weights. However, as in traditional deep learning, setting all means to zero prevents messages from breaking symmetry. To resolve this, we sample prior means according to spectral parametrization \citep{yang2024spectralconditionfeaturelearning}, which facilitates feature learning independent of the network width. Another challenge in prior choice is managing exploding variances. In a naive setup with $\sigma^2_p = 1$, forward message variances grow exponentially with the network depth. While we attempted to find a principled choice of $\sigma^2_p$, our current initialization scheme is based on experimental data (see \ref{appendix:prior_analysis}). For a layer with $d_1$ inputs and $d_2$ outputs, we set
$$
\sigma^2_p = \frac{1.5 - 0.8041 \cdot \min(1.0, d_2 / d_1)}{0.8041 + 0.4496 \cdot d_1}.
$$
Refer to \ref{appendix:prior_analysis} for our justification of this formula.

% Should we mention orthogonal initalization? Do we have any experiments that use it?

\section{Experiments}
% Some quote that was in this section a bunch of times: Focus on (1) accuracy, (2) learning curve, and (3) calibration/rejection

\subsection{Synthetic Data} % Romeo
% 1. Depth Scaling
% - Setup: Hidden layers of width 16 with increasing depth. 200 datapoints in a sine curve dataset over 2 periods.
% - The results at a depth of 4-5 are much better than at 2-3 (not expressive enough to fit this data) or at 6+ (too deep to fit well; similar problem can be observed for Torch models)
% - At depth 4-5, our model fits this complicated dataset well.
% - The behavior at the edges is linear, both for the mean prediction as well as the standard deviations. This makes sense given our linear model

% 2. Out-of-distribution experiment (sine curves with data in (-0.5, +0.5)). This can be running text, no need to spend half a page on the chart
% - Goal of the experiment: Figure out how well the coverage probabilities work for areas outside of the data.
% - Setup: Same sine curves, but over a smaller area (only -0.5 to +0.5)

\begin{figure}[htbp]
  \centering
  \begin{subfigure}[b]{0.245\textwidth}
    \includegraphics[width=\textwidth]{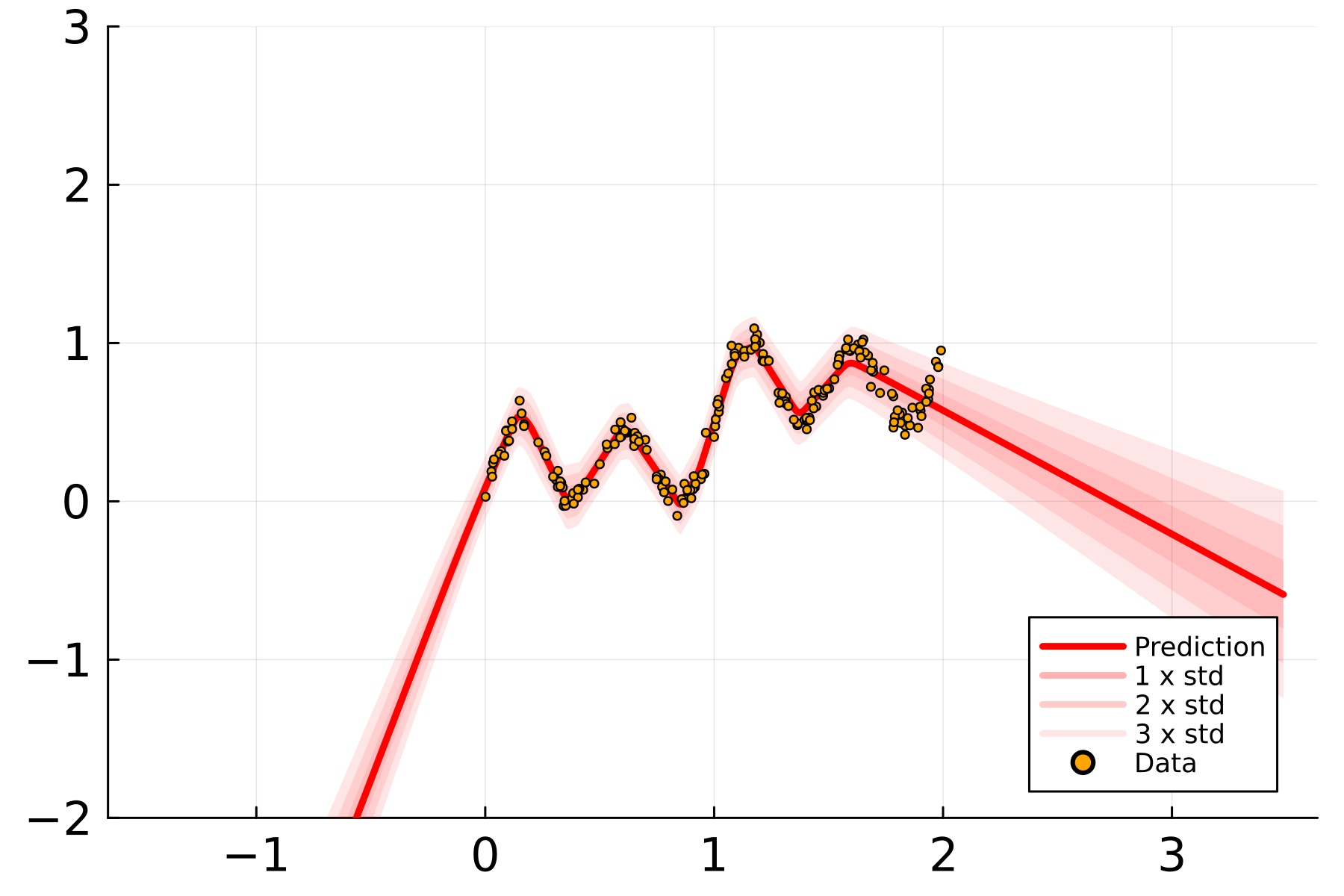}
    \caption{Three layers}
  \end{subfigure}
  \begin{subfigure}[b]{0.245\textwidth}
    \includegraphics[width=\textwidth]{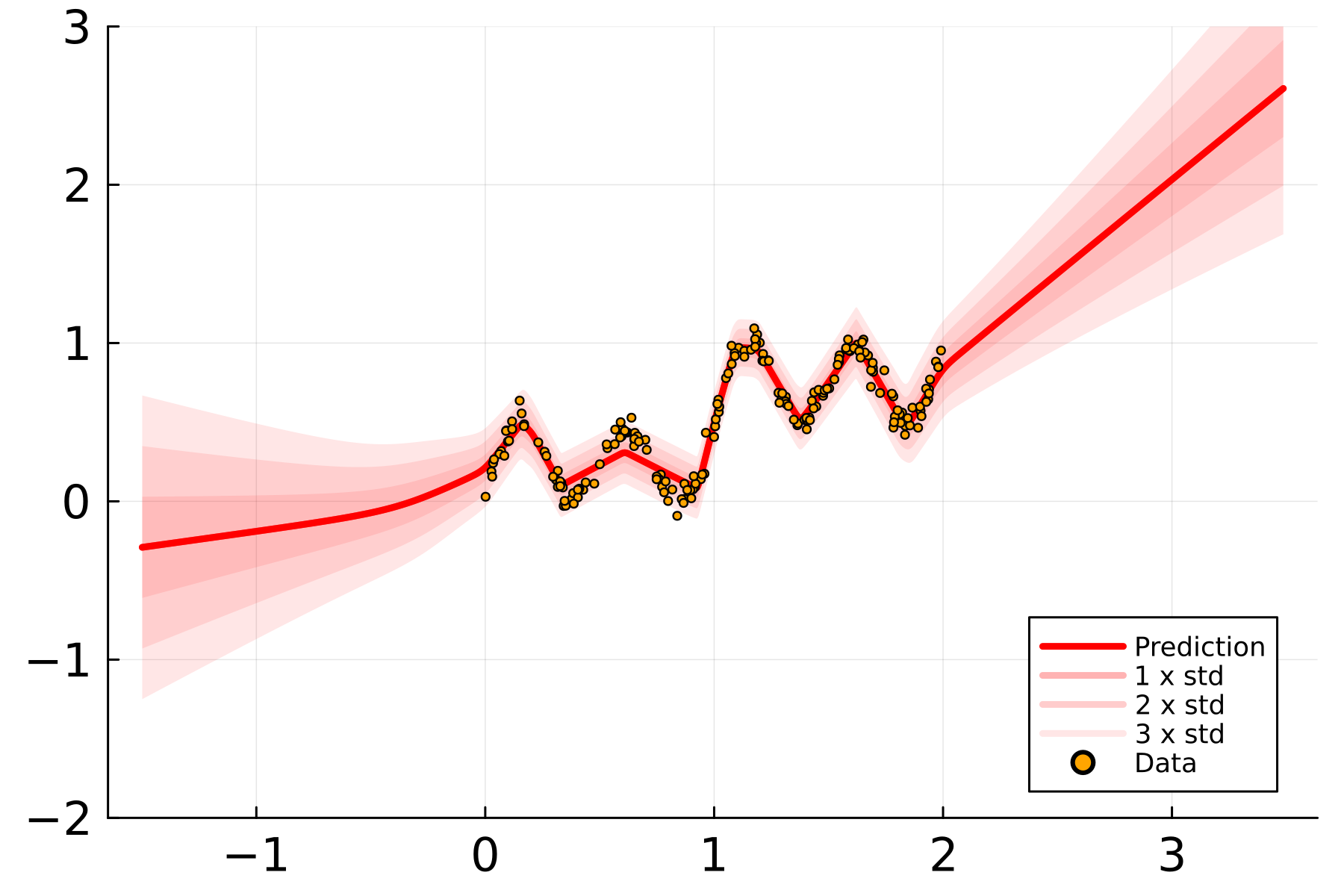}
    \caption{Four layers}
  \end{subfigure}
  \begin{subfigure}[b]{0.245\textwidth}
    \includegraphics[width=\textwidth]{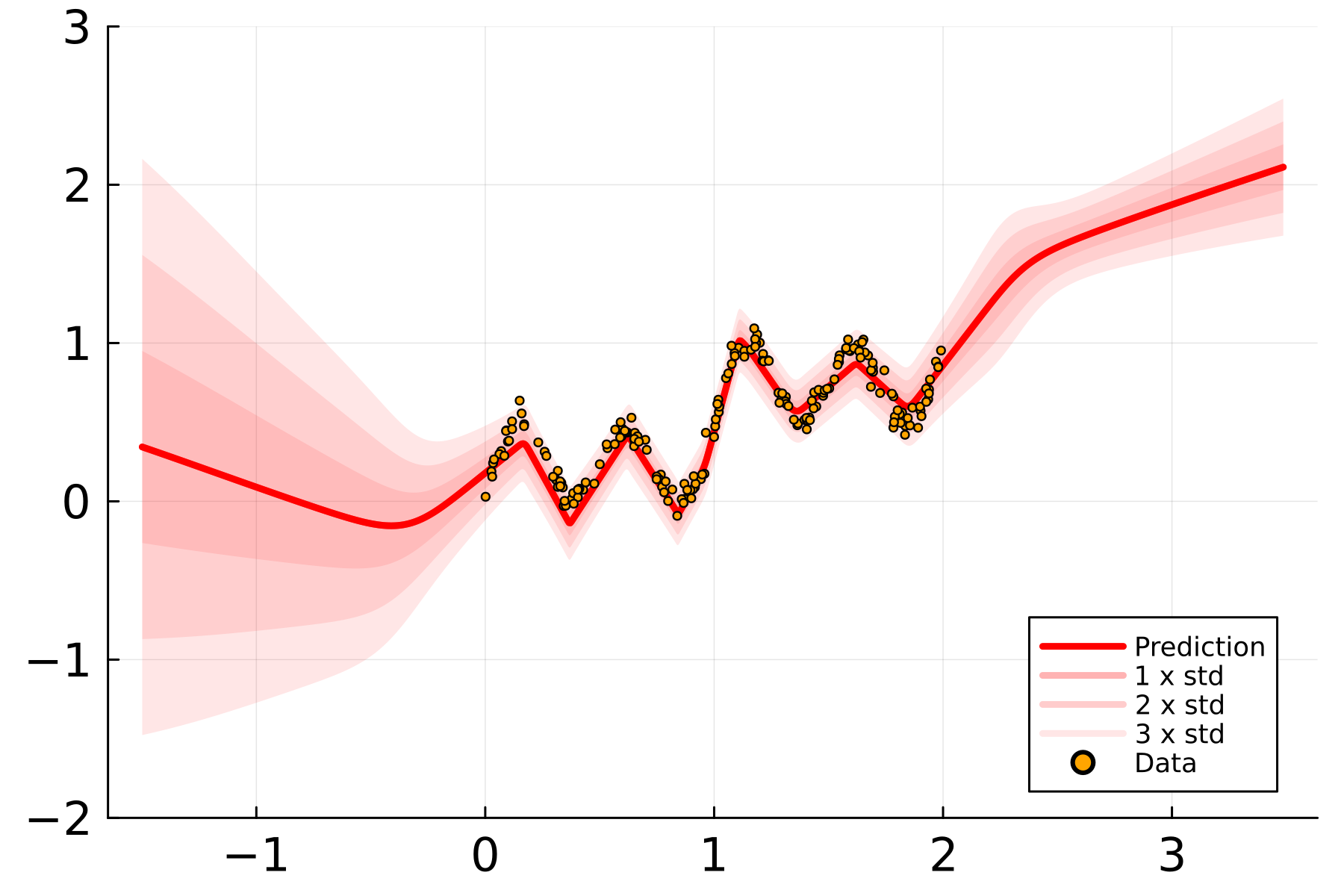}
    \caption{Five layers}
  \end{subfigure}
  \begin{subfigure}[b]{0.245\textwidth}
    \includegraphics[width=\textwidth]{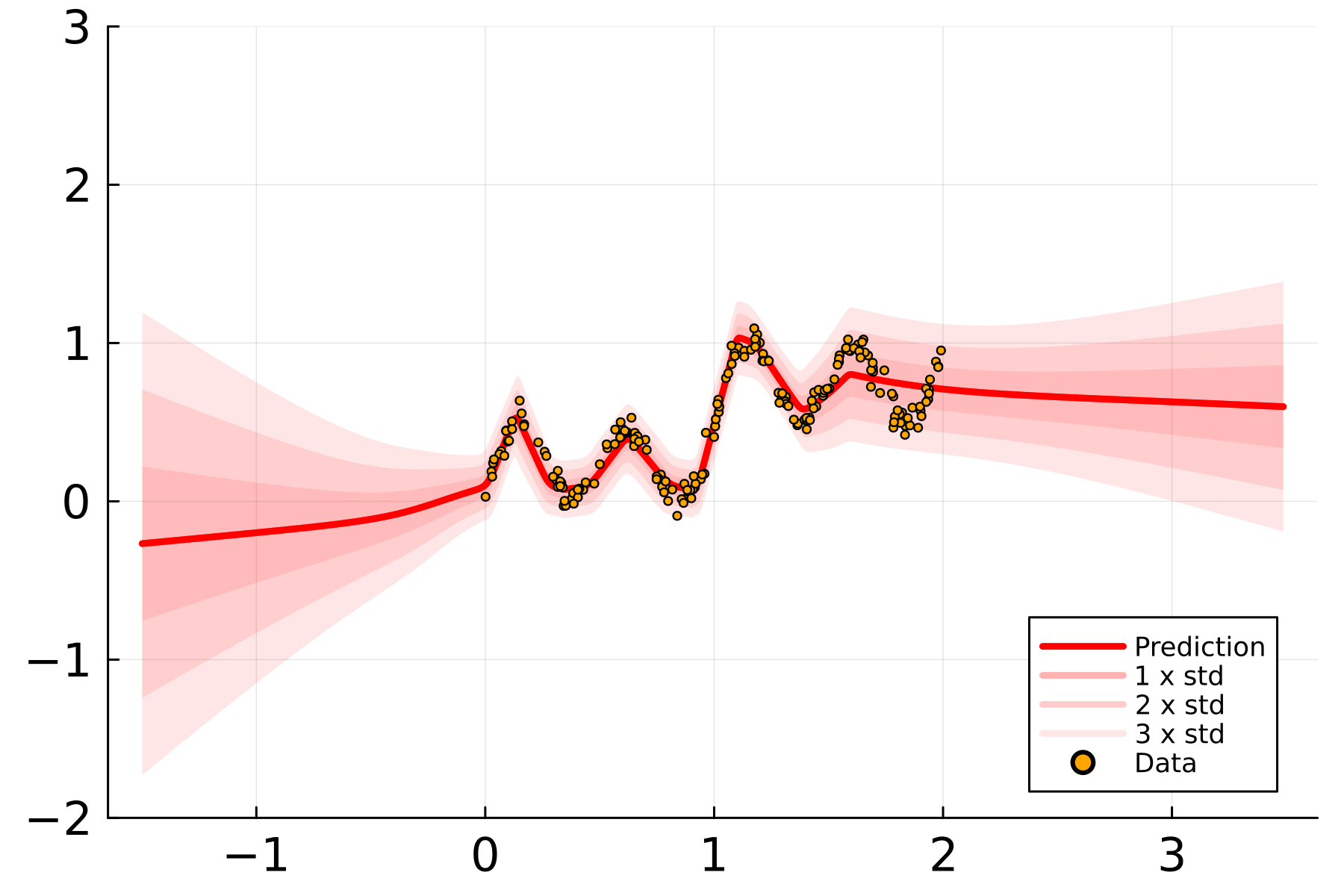}
    \caption{Six layers}
  \end{subfigure}

  \caption{Fitting MLPs of width 16 with increasing depth. Between any linear layers we apply LeakyReLU with a leak of $0.1$. As the depth increases, the network becomes more expressive but harder to fit.}
  \label{fig:depth_scaling}
\end{figure}

We first evaluate our model on a synthetic sine curve dataset of 200 data points. \Cref{fig:depth_scaling} shows that an MLP with 4-5 linear layers fits the data well, whereas smaller models are not expressive enough to capture the data and deeper models are harder to fit. For depths beyond six layers, the performance degrades further, but the same is true for models of the same architecture trained in Torch. As expected, the posterior approximations in \Cref{fig:depth_scaling} have small variance within the training range and high variance outside or when the fit is bad. In all plots, the mean prediction and standard deviation expand linearly outside the training range.

To assess how well our model's posterior uncertainty generalizes beyond the training data, we trained 100 separate models on the same sine curve data and evaluated their performance on unseen inputs. For this test, we limit the training data range to $(-0.5, 0.5)$ and then measure if the posterior approximation covers the true data-generating function outside of this training range. For negative $x$, $61\%$ of $1\sigma$-intervals covered the true data-generating function, $86\%$ of $2\sigma$-intervals, and $93\%$ of $3\sigma$-intervals. For positive $x$, we measured $36\%$, $68\%$, and $90\%$ respectively. While these measurements are slightly lower than the probability mass covered by the respective intervals, the posterior uncertainty appears to be reasonably well-calibrated. Overall, we found a strong correlation of 0.90 between credible intervals of the predictive posterior and the coverage rate.

\subsection{CIFAR-10}
To evaluate our method on the CIFAR-10 dataset we trained a 6 layer deep convolutional network with roughly 890k parameters on the full training dataset.
As baseline methods we picked the SOTA optimizers AdamW \citep{AdamW} and IVON \citep{ivon_shen2024variationallearningeffectivelarge} each with a cosine annealing learning rate schedule \citep{CosineAnnealing}.
Across all methods, including ours, we trained for 25 epochs.
In \Cref{appendix:experimental_setup} we give extensive details on the network architecture and the experimental setup in general.
\Cref{tab:cifar-10_experiments} compares the performance of our method (MP) against AdamW  and IVON  across a variety of standard metrics.
In general, we see that MP can compete with these two strong baselines.
And in the expected calibration error our method even has a notable edge.
That the metrics are overall worse than what is reported by \citet{ivon_shen2024variationallearningeffectivelarge} is likely due to a difference in architecture; Shen et al. only conduct experiments on ResNets equipped with filter response normalization \citep{Filter_Response_Normalization}.
Neither residual connections nor normalization layers are yet implemented in our factor graph library.
Nevertheless, these results motivate to further improve our approach.
In the future work part of \Cref{section:conlusion_and_future_work} we outline ideas on how to model such factors.

\begin{table}[h]
\centering
\begin{tabular}{@{}lcccccc@{}}
\toprule
        & Acc. $\uparrow$ & Top-5 Acc. $\uparrow$ & NLL $\downarrow$ & ECE $\downarrow$ & Brier $\downarrow$ & OOD-AUROC $\uparrow$ \\ \midrule
AdamW	& \textbf{0.783} & \textbf{0.984} & 1.736 & 0.046 & 0.38 & 0.792 \\
IVON@mean	& 0.772 & 0.983 & 1.494 & 0.041 & 0.387 & \textbf{0.819} \\
IVON	& 0.772 & 0.983 & 1.316 & 0.035 & 0.37 & 0.808 \\
MP (Ours) & 0.773 & 0.977 & \textbf{0.997} & \textbf{0.029} & \textbf{0.361} & 0.81 \\
\bottomrule
\end{tabular}
\caption{Comparison of various validation statistics for a convolutional network of roughly 890k parameters trained on CIFAR-10. Out-of-distribution (OOD) detection was tested with SVHN. For IVON we used 100 samples for prediction at test time. IVON@mean are the results obtained from evaluating the model at the means of the learned distributions of the individual parameters.}
\label{tab:cifar-10_experiments}
\end{table}
\textbf{Reproducibility}
All code is available at \href{https://github.com/christian-helms/mpbnns.git}{https://github.com/christian-helms/mpbnns.git}.
% There are several factors to consider when comparing the runtime of our MP system to Torch (SGD). First, our message equations involve more complex computations than basic matrix multiplications, which adds inherent overhead. However, additional factors also contribute to the current performance difference. Our implementation relies on Tullio for CUDA kernel generation and does not yet incorporate memory-layout optimization or other hand-optimized features seen in mature libraries like Torch. The absence of batch computation further limits training efficiency. Another consideration is that our implementation currently relies on Julia’s default FP64 precision, which the GPU processes at a lower throughput (1:32 of FP32 \citep{teslaT4}). Despite these implementation-related details, the overall training runtime scales linearly with the number of weights and examples.

\section{Conclusion}
\label{section:conlusion_and_future_work}

\textbf{Summary:} We presented a novel framework that advances message-passing (MP) for Bayesian neural networks by modeling the predictive posterior as a factor graph. To the best of our knowledge, this is the first MP method to handle convolutional neural networks while avoiding double-counting training data, a limitation in previous MP approaches like \citet{expectation_backpropagation,hernándezlobato2015probabilisticbackpropagationscalablelearning,Lucibello_2022}. 
In our experiment on the CIFAR-10 dataset our method proofed to be competitive with the SOTA baselines AdamW and IVON, even showing an edge in terms of calibration. 

\textbf{Limitations: }
Despite recent advances, variational inference methods like IVON remain ahead in scale and performance on larger datasets. Our approach's runtime and memory requirements scale linearly with model parameters and dataset size. While our inference at test time can keep up with IVON's sampling approach in terms of speed and memory requirements, training is up to two orders of magnitude slower and more GPU-memory intensive compared to training deterministic networks using PyTorch with optimizers like AdamW.

The memory overhead stems from two key factors: First, each training example stores messages proportional to the model's parameter count, unlike AdamW's batch-level intermediate representations. Second, each parameter requires two 8-byte floating-point numbers, contrasting with more efficient 4-byte or smaller formats.

Runtime inflation results from several performance bottlenecks: Our training schedule lacks parallel forward passes, our Tullio-based CUDA kernel generation misses memory-layout and GPU optimizations present in mature libraries like Torch, message equations involve complex computations beyond standard matrix multiplications, and we use Julia's default FP64 precision, which GPUs process less efficiently.

\textbf{Future Work: }
We believe Moment Propagation (MP) holds significant promise for more balanced uncertainty estimates, thanks to its moment-matching ability, compared to Variational Inference's tendency toward overconfident predictions. Further improvements in scalability and architectural flexibility could make MP a competitive alternative to VI.

Concretely, in terms of memory requirements, it is worth exploring whether iterating on individual examples instead of batches, and starting from scratch in each epoch, could leave our method ahead.
While this might reintroduce the double counting problem, it would drastically reduce the GPU-memory footprint.
Regarding training efficiency, an altered message-update schedule with actual batched computations would significantly reduce training time. Reimplementing our library in CUDA C++ with efficiency in mind could also drastically cut down computational overhead.

On the architectural front, we deem it likely that our approach can be extended to most modern deep learning architectures. Residual connections are straightforward to implement as they boil down to simple sum factors. For normalization layers at the scalar level, only a division factor is missing, which can be approximated by a "rotated" product factor. This would suffice to model ResNet-like architectures and more modern convolutional networks like ConvNeXt.
For transformers, the last ingredient needed is an efficient softargmax factor. Given the division factor, only an exp factor is missing to model softargmax at the scalar level.

Finally, future work might also explore applications to more applied tasks such as continual learning, sparse networks, and Bayesian reinforcement learning.

\bibliography{main}
\bibliographystyle{template}

\appendix
\section{Proof of Global Minimization Objective}

\subsection{Moment-Matched Gaussians Minimize Cross-Entropy}
\label{appendix:moment_match_minimization}
Consider a scalar density $p$ and a Gaussian $q(\theta) = \N(\theta, \mu,\sigma)$.
Then
\begin{align*}
    \min H(p, q) = \min \left( \int p(\theta) \log\left(\frac{p(\theta)}{q(\theta)}\right) d\theta \right) = \min \left( \frac{1}{2\sigma^2} \int p(\theta)(\theta - \mu)^2\,d\theta + \frac{\log(2\pi \sigma^2)}{2} \right).
\end{align*}
It is well known that expectations minimize the expected mean squared error. In other words, the integral is minimized by setting $\mu$ to the expectation of $p$ and is then equal to the variance of $p$.
The necessary condition of a local minimum then yields that $\sigma^2$ must be the variance of $p$.

\subsection{Proof of \texorpdfstring{\Cref{equation:posterior_approx_minimization}}: Global Minimization Objective}
\label{appendix:minimization_proof}
Let $p$ be an arbitrary probability density on $\R^k$ with marginals $p_i(\theta_i) := \int p(\theta)\,d(\theta \setminus \theta_i)$ and denote by $\mathcal{Q}$ the set of diagonal Gaussians.
Then for every $q(\vtheta) = \prod_{i=1}^k q_i(\theta_i) \in\mathcal{Q}$ we can write the relative entropy from $p$ to $q$ as
\begin{align*}
    \KL{[\,p\,||\,q\,]} &= \int p(\vtheta)\log\left(\frac{p(\vtheta)}{q(\vtheta)}\right)d\vtheta = - \sum_{i=1}^k \int p(\vtheta)\log(q(\theta_i))d\vtheta - H(p)\\
    &= -\sum_{i=1}^k \int_{\theta_i} \log(q_i(\theta_i)) \int_{\vtheta \setminus \theta_i}p(\vtheta)d(\vtheta \setminus \theta_i) - H(p) = \sum_{i=1}^k H(p_i, q_i) - H(p).
\end{align*}
This shows that $\KL{[\,p\,||\,q\,]}$ is minimized by independently minimizing the summands $H(p_i,q_i).$
In combination with \ref{appendix:moment_match_minimization} this completes the proof. 

\section{Derivations of Message Equations}

\subsection{ReLU}
A common activation function is the Rectified Linear Unit $\text{ReLU}:\R\rightarrow\R,z\mapsto \text{max}(0,z).$
\paragraph{Forward Message:} Since ReLU is not injective, we cannot apply the density transformation property of the Dirac delta to the forward message
$$\msg{f}{\ra}(\ra) = \int_{\rz\in\R}\delta(\ra-\text{ReLU}(\rz))\msg{\rz}{f}(\rz)\,d\rz.$$
In fact, the random variable $\text{ReLU}(\rZ)$ with $\rZ \sim \msg{z}{f}$ does not even have a density.
A positive amount of weight, namely $\Pr[Z \leq 0]$, is mapped to 0.
Therefore
$$\msg{f}{\ra}(0) = \lim_{t \rightarrow 0}\int_{\rz\in\R}\N(\text{ReLU}(\rz);0,t^2)\msg{\rz}{f}(\rz)\,d\rz \geq \lim_{t \rightarrow 0} \N(0; 0, t^2)\min_{\rz\in[-1,0]}\msg{\rz}{f}(\rz) = \infty.$$
Apart from $0$, the forward message is well defined everywhere, and technically null sets do not matter under the integral.
However, moment-matching $\msg{\rz}{f}$ while truncating at 0 does not seem reasonable as it completely ignores the weight of $\msg{\rz}{f}$ on $\R_{\leq 0}$.
Therefore, we refrain from moment-matching the forward message of ReLU.
   
As an alternative, we consider a marginal approximation.
That means, we derive formulas for
\begin{equation}
    \label{equation:marginal_approximation_moments}
    m_k := \int_{\ra\in\R} \ra^k\msg{\ra}{f}(\ra)\msg{f}{\ra}(\ra)\,d\ra,~~~k\in\{0,1,2\}
\end{equation}
and then set
$$\msg{f}{\ra}(\ra) := \N(\ra; m_1 / m_0, m_2 / m_0 - (m_1 / m_0)^2)\,/\,\msg{\ra}{f}(\ra).$$
By changing the integration order, we obtain
\begin{align*}
    m_k &= \int_{\ra\in\R}\ra^k \msg{\ra}{f}(\ra)\int_{z\in\R}\delta(\ra - \text{ReLU}(z))\msg{z}{f}(z)\,dz\,d\ra \\
        &= \int_{z\in\R}\msg{z}{f}(z)\int_{\ra\in\R}\delta(\ra - \text{ReLU}(z)) \ra^k\msg{\ra}{f}(\ra)\,d\ra\,dz \\
        &= \int_{z\in\R}\msg{z}{f}(z) \text{ReLU}^k(z)\msg{\ra}{f}(\text{ReLU}(z))\,dz \\
\end{align*}
Note that we end up with a well-defined and finite integral.
Similar integrals arise in later derivations.
For this reason we encapsulate part of the analysis in basic building blocks.
\begin{building_block}
    \label{building_block:product_of_gaussians_positive}
    We can efficiently approximate integrals of the form
    $$\int_0^\infty z^k\N(z;\mu_1,\sigma^2_1)\N(z; \mu_2, \sigma^2_2)\,dz$$
    where $\mu_1,\mu_2\in\R, \sigma_1,\sigma_2 > 0$ and $k=0,1,2$.
\end{building_block}
\begin{proof}
    By \Cref{equation:multiplying_natural_parameter_gaussians} the integral is equal to 
\begin{align*}
    S^+ &= \N(\mu_1; \mu_2, \sigma^2_1 + \sigma^2_2)\int_0^\infty z^k\N\left(z;\mu,\sigma^2\right)\,dz\\
        &= \N(\mu_1; \mu_2, \sigma^2_1 + \sigma^2_2) \begin{cases}
            \E[\text{ReLU}^k(\N(\mu, \sigma^2))]~~~&\text{for }k=1,2 \\
            \Pr[-Z \leq 0] = \phi(\mu / \sigma)      ~~~&\text{for }k=0
        \end{cases}
\end{align*}
where
$$\mu = \frac{\tau}{\rho},~~~\sigma^2 = \frac{1}{\rho},~~~\tau = \frac{\mu_1}{\sigma^2_1} + \frac{\mu_2}{\sigma^2_2}~~~\text{and}~~~\rho = \frac{1}{\sigma^2_1} + \frac{1}{\sigma^2_2}.$$
\end{proof}

This motivates the derivation of efficient formulas for the moments of an image of a Gaussian variable under ReLU. 
\begin{building_block}
\label{appendix:building_block_relu_moments}
Let $\rZ \sim \N(\mu, \sigma^2)$.
The first two moments of $\text{ReLU}(\rZ)$ are then given by
\begin{align}
    \E[\text{ReLU}(\rZ)] &= \sigma \varphi(x) + \mu \phi(x)\\
    \E[\text{ReLU}^2(\rZ)] &= \sigma\mu \varphi(x) + (\sigma^2 + \mu^2)\phi(x),
\end{align}
where $x = \mu / \sigma$ and $\varphi,\phi$ denote the pdf and cdf of the standard normal distribution, respectively.
\end{building_block}
\begin{proof}
    
The basic idea is to apply $\int ze^{-z^2/2}\,dz = -e^{-z^2/2}.$ 
Together with a productive zero, one obtains
    \begin{align*}
        \sqrt{2\pi}\sigma\E[\text{ReLU}(\rZ)] &= \int_0^\infty z e^{-\frac{(z-\mu)^2}{2\sigma^2}}\,dz =  \sigma^2\int_0^\infty \frac{(z-\mu)}{\sigma^2} e^{-\frac{(z-\mu)^2}{2\sigma^2}}\,dz + \mu \int_0^\infty e^{-\frac{(z-\mu)^2}{2\sigma^2}}\,dz \\
        &= \sigma^2 \left[-e^{-\frac{(z-\mu)^2}{2\sigma^2}}\right]^\infty_0 + \sqrt{2\pi}\sigma\mu\Pr[Z \geq 0] \\
        &= \sigma^2 e^{-\frac{\mu^2}{2\sigma^2}} + \sqrt{2\pi}\sigma\mu \Pr\left[\frac{-Z + \mu}{\sigma} \leq \frac{\mu}{\sigma} \right] \\
        &= \sqrt{2\pi}\sigma^2 \varphi(x) + \sqrt{2\pi}\sigma\mu \phi(x).
    \end{align*}
    Rearranging yields the desired formula for the first moment.
    For the second moment, we need to complete the square and perform integration by parts:
    \begin{align*}
        &~~~~\E[\text{ReLU}^2(\rZ)] = \frac{1}{\sqrt{2\pi}\sigma}\int_0^\infty z^2 e^{-\frac{(z-\mu)^2}{2\sigma^2}}\,dz \\
        &= \frac{1}{\sqrt{2\pi}\sigma}\left(\sigma^2\int_0^\infty (z-\mu)\frac{z-\mu}{\sigma^2}e^{-\frac{(z-\mu)^2}{2\sigma^2}}\,dz + 2\mu\int_0^\infty ze^{-\frac{(z-\mu)^2}{2\sigma^2}}\,dz - \mu^2\int_0^\infty e^{-\frac{(z-\mu)^2}{2\sigma^2}}\,dz\right) \\
        &= \frac{\sigma^2}{\sqrt{2\pi}\sigma} \left(\left[-(z-\mu)e^{-\frac{(z-\mu)^2}{2\sigma^2}}\right]^\infty_0 + \int_0^\infty e^{-\frac{(z-\mu)^2}{2\sigma^2}}\right) + 2\mu\E[\text{ReLU}(Z)] - \mu^2 \phi(x) \\
        &= -\sigma\mu\varphi(x) + \sigma^2\phi(x) + 2\mu\E[\text{ReLU}(\rZ)] - \mu^2 \phi(x)
        = \sigma\mu \varphi(x) + (\sigma^2 + \mu^2)\phi(x).
    \end{align*}
\end{proof}

\begin{building_block}
    \label{building_block:product_of_gaussians_with_relu_negative}
    Integrals of the form
    $$S^- := \int_{-\infty}^0 z^k\N(z;\mu_1,\sigma^2_1)\N(0; \mu_2, \sigma^2_2)\,dz$$
    where $\mu_1,\mu_2\in\R, \sigma_1,\sigma_2 > 0$ and $k=0,1,2$ can be efficiently approximated.
\end{building_block}
\begin{proof}
    Employing the substitution $z = -t$ gives
\begin{align*}
    S^- &= \N(0;\mu_2,\sigma^2_2) \int_0^\infty (-1)^kt^k \N(-t; \mu_1,\sigma^2_1)\,dt = (-1)^k\N(0;\mu_2,\sigma^2_2)\int_0^\infty t^k\N(t; -\mu_1,\sigma^2_1)\,dt\\
        &= (-1)^k\N(0;\mu_2,\sigma^2_2) \begin{cases}
            \E[\text{ReLU}(\N(-\mu_1,\sigma^2_1))]~~~&\text{for }k=1,2 \\
            \Pr[-Z \geq 0] = \phi(-\mu_1 / \sigma_1)~~~&\text{for }k=0.
        \end{cases}
\end{align*}
\end{proof}

Now let $\msg{z}{f}(z) = \N(z;\mu_z,\sigma_z^2), \msg{\ra}{f}(\ra)=\N(\ra;\mu_\ra,\sigma^2_\ra)$ and consider the decomposition
$$m_k = \underbrace{\int_0^\infty z^k\N(z;\mu_z,\sigma^2_z)\N(z; \mu_\ra, \sigma^2_\ra)\,dz}_{S^+} + \underbrace{\int_{-\infty}^0 \text{ReLU}^k(z)\N(z; \mu_z, \sigma^2_z)\N(0; \mu_\ra, 
\sigma^2_\ra)\,dz}_{S^-}.$$
Note that $S^+$ falls under Building Block \ref{building_block:product_of_gaussians_positive} for any $k=0,1,2.$
The other addend $S^-$ is equal to 0 for $k=1,2$, and is handled by Building Block \ref{building_block:product_of_gaussians_with_relu_negative} for $k=0$. 

\paragraph{Backward Message:}
By definition of the Dirac delta, the backward message is equal to
$$\msg{f}{z}(z) = \int_{\ra\in\R}\delta(\ra - \text{ReLU}(z))\msg{\ra}{f}(\ra)\,d\ra = \msg{\ra}{f}(\text{ReLU}(z))$$
which is, of course, not integrable, so it cannot be interpreted as a scaled density.
Instead, we apply marginal approximation by deriving formulas for
$$m_k := \int_{z\in\R} z^k\msg{z}{f}(z)\msg{f}{z}(z)\,dz,~~~k\in\{0,1,2\}$$
and then setting
$$\msg{f}{z}(z) := \N(z; m_1 / m_0, m_2 / m_0 - (m_1 / m_0)^2)\,/\,\msg{z}{f}(z).$$
To this end, let $\msg{z}{f}(z) = \N(z;\mu_z,\sigma^2_z)$ and $\msg{\ra}{f}(\ra) = \N(\ra; \mu_\ra,\sigma^2_\ra)$.
Then we have
$$m_k = \underbrace{\int_0^\infty z^k\N(z;\mu_z,\sigma^2_z)\N(z; \mu_\ra, \sigma^2_a)\,dz}_{S^+} + \underbrace{\int_{-\infty}^0 z^k\N(z; \mu_z, \sigma^2_z)\N(0; \mu_a, 
\sigma^2_a)\,dz}_{S^-}.$$
The two addends $S^+$ and $S^-$ are handled by Building Block \ref{building_block:product_of_gaussians_positive} and Building Block \ref{building_block:product_of_gaussians_with_relu_negative}, respectively.

\subsection{Leaky ReLU}
\label{appendix:LeakyReLU_derivations}
\begin{figure}[ht!]
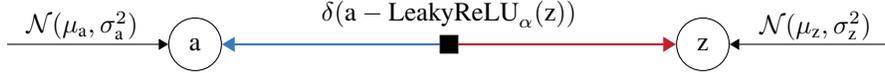

    \centering
    \ctikzfig{leaky_relu}
    \caption{A deterministic factor corresponding to the $\text{LeakyReLU}_\alpha$ activation function.}
    \label{fig:leaky_relu_factor}
\end{figure}

Another common activation function is the Leaky Rectified Linear Unit
$$\text{LeakyReLU}_\alpha:\R\rightarrow\R,z\mapsto \begin{cases}
    z~~~&\text{for }z\geq 0\\
    \alpha z ~~~&\text{for }z < 0.
\end{cases}$$
It is parameterized by some $\alpha > 0$ that is typically small, such as $\alpha = 0.1$.
In contrast to ReLU, it is injective (and even bijective).
For this reason the forward and backward messages are both integrable and can be approximated by both direct and marginal moment matching.
The notation is shown in \Cref{fig:leaky_relu_factor}.

\paragraph{Forward Message:} It is easy to show that the density of $\text{LeakyReLU}_\alpha(\N(\mu_z,\sigma^2_z))$ is given by
$$p(a) = \N(\text{LeakyReLU}_{1/\alpha}(a);\mu_z,\sigma^2_z) \begin{cases}
    1~&\text{for } z \geq 0\\
    1/\alpha~&\text{for } z < 0
\end{cases}$$
which only has one discontinuity point, namely 0.
In particular, it is continuous almost everywhere.
% So by Proposition \ref{proposition:delta_density_transformation}, ...
So by the density transformation property of Dirac's delta, we have $\msg{f}{a}(a) = p(a)$ for almost all $a$.
Under the integral we can therefore replace $\msg{f}{a}(a)$ by $p(a)$.
This justifies that the moments of $\msg{f}{a}$ are exactly the moments of $(\text{LeakyReLU}_\alpha)_*\N(\mu_z,\sigma^2_z)$. 
Its expectation is equal to
\begin{align*}
    \E\,[\text{LeakyReLU}_\alpha(\N(\mu_z,\sigma^2_z))] &= \int_{-\infty}^0 \alpha z\N(z;\mu_z,\sigma_z^2)\,dz+ \int_0^\infty z\N(z;\mu_z,\sigma_z^2)\,dz \\
    &= -\alpha \int_0^\infty t\N(t;-\mu_z,\sigma^2_z)\,dt + \int_0^\infty z\N(z;\mu_z,\sigma_z^2)\,dz \\
    &= -\alpha \E[\text{ReLU}(\N(-\mu_z,\sigma^2_z))] + \E[\text{ReLU}(Z)].
\end{align*}
Both addends are handled by Building Block \ref{appendix:building_block_relu_moments}.
Yet we can get more insight by further substitution:
\begin{align*}
    \E[\text{LeakyReLU}_\alpha(Z)] &= -\alpha (\sigma_z\varphi(-\mu_z/\sigma_z) - \mu_z \phi(-\mu_z / \sigma_z)) + \sigma_z\varphi(\mu_z / \sigma_z) + \mu_z\phi(\mu_z / \sigma_z) \\
    &= (1 - \alpha)(\sigma_z\varphi(\mu_z /\sigma_z) + \mu_z\phi(\mu_z / \sigma_z)) + \alpha \mu_z \\
    &= (1 - \alpha)\E[\text{ReLU}(Z)] + \alpha \E[Z].
\end{align*}
In the second to last equation, we use the identities $\varphi(-x) = \varphi(x)$ and $\phi(-x) = 1 - \phi(x).$
As such, the mean of $\text{LeakyReLU}_\alpha(Z)$ is a convex combination of the mean of $\text{ReLU}(Z)$ and the mean of $Z$.
The function $\text{LeakyReLU}_1$ the identity, and its mean is accordingly the mean of $Z$.
For $\alpha = 0$, we recover the mean of $\text{ReLU}(Z).$

The second moment of $\text{LeakyReLU}_\alpha(Z)$ decomposes to
\begin{align*}
    \E[\text{LeakyReLU}^2_\alpha(Z)] &= \int_{-\infty}^0 \alpha^2 z^2 \N(z;\mu_z, \sigma_z^2)\,dz + \int_0^\infty z^2 \N(z;\mu_z,\sigma_z^2)\,dz \\
    &= \alpha^2 \int_0^\infty z^2\N(z;-\mu_z,\sigma_z^2)\,dz + \int_0^\infty z^2\N(z;\mu_z,\sigma^2_z)\,dz \\
    &= \alpha^2 \E[\text{ReLU}^2(\N(-\mu_z,\sigma_z^2))] + \E[\text{ReLU}^2(\N(\mu_z,\sigma^2_z))].
\end{align*}
Again, both addends are covered by Building Block \ref{appendix:building_block_relu_moments}, so approximating the forward message via direct moment matching is feasible.

A marginal approximation can also be found.
For all $k=0,1,2$ we have
\begin{align*}
    &\int_{a\in\R}a^k \msg{a}{f}(a)\msg{f}{a}(a)\,da = \int_{a\in\R} a^k\msg{a}{f}(a)p(a)\,da\\
    = &\underbrace{\frac{1}{\alpha}\int_{-\infty}^0 a^k\N(a; \mu_a, \sigma^2_a)\N(a/\alpha; \mu_z,\sigma^2_z)\,da}_{S^-} + \underbrace{\int_0^\infty a^k\N(a; \mu_a, \sigma^2_a)\N(a; \mu_z,\sigma^2_z)\,da}_{S^+}
\end{align*}
The term $S^+$ is handled by Building Block \ref{building_block:product_of_gaussians_positive}.
The term $S^-$ is equal to
\begin{align*}
    S^- &= \int_{-\infty}^0 a^k\N(a; \mu_a, \sigma^2_a)\N(a; \alpha\mu_z,(\alpha\sigma_z)^2)\,da \\
    &= (-1)^k \int_0^\infty a^k \N(a; -\mu_a, \sigma^2_a)\N(a; -\alpha\mu_z,(\alpha\sigma_z)^2)\,da 
\end{align*}
and therefore also covered by Building Block \ref{building_block:product_of_gaussians_positive}.

\paragraph{Backward Message:} By the sifting property of the Dirac delta, the backward message is equal to
$$\msg{f}{z}(z) = \int_{a\in\R}\delta(a - \text{LeakyReLU}_\alpha(z))\msg{a}{f}(a)\,da = \msg{a}{f}(\text{LeakyReLU}_\alpha(z)).$$
As opposed to ReLU, the backward message is integrable.
That means, we can also apply direct moment matching:
For all $k=0,1,2$ we have
\begin{align*}
    \msg{f}{z}(z) &= \int_{-\infty}^0 z^k \N(\alpha z; \mu_a,\sigma_a^2)\,dz + \int_{0}^\infty z^k \N(z; \mu_a,\sigma_a^2)\,dz\\
                  &= \frac{(-1)^k}{\alpha} \int_{0}^\infty z^k \N(z; -\mu_a/\alpha,(\sigma_a/\alpha)^2)\,dz + \int_{0}^\infty z^k \N(z; \mu_a,\sigma_a^2)\,dz
\end{align*}
For $k=1$ or $k=2$, the integrals fall under Building Block \ref{appendix:building_block_relu_moments} again.
If $k=0$, then 
$$\msg{f}{z}(z) = \frac{(-1)^k}{\alpha} \phi(-\mu_a / \sigma_a) + \phi(\mu_a / \sigma_a).$$

Again, we can also find a marginal approximation as well.
For all $k=0,1,2$, we can write
\begin{align*}
    &\int_{z\in\R} z^k\msg{z}{f}(z)\msg{f}{z}(z)\,dz \\
    = &\int_{-\infty}^0 z^k \N(z; \mu_z,\sigma_z^2)\N(\alpha z; \mu_a,\sigma_a^2) \,dz + \int_0^\infty z^k \N(z; \mu_z,\sigma_z^2)\N(z; \mu_a,\sigma_a^2)\,dz \\
    = & \frac{(-1)^k}{\alpha}\int_{0}^\infty z^k \N(z; -\mu_z,\sigma_z^2)\N(z; -\mu_a/\alpha,(\sigma_a/\alpha)^2) \,dz + \int_0^\infty z^k \N(z; \mu_z,\sigma_z^2)\N(z; \mu_a,\sigma_a^2)\,dz 
\end{align*}
Since both integrals are covered by Building Block \ref{building_block:product_of_gaussians_positive} we have derived direct and marginal approximations of LeakyReLU messages using moment matching.

\subsection{Softmax}

We model the soft(arg)max training signal as depicted in \Cref{tab:message_equations_training_signals}.
For the forward message on the prediction branch, we employ the so-called "probit approximation" \citep{daxberger2022laplacereduxeffortless}:
$$\msg{f}{\rc}(i) = \int \text{softmax}(\rva)_i \N(\rva; \boldsymbol{\mu},\text{diag}(\boldsymbol{\sigma}^2)\,d\rva \approx \text{softmax}(\vt)_i,$$
where $t_j = \mu_j/ (1 + \frac{\pi}{8}\sigma_j^2),~j=1,\dots,d$.
For the backward message on a training branch, to say $\ra_d$, we use marginal approximation. We hence need to compute the moments $m_0,m_1,m_2$ of the marginal of $\ra_d$ via:
\begin{align*}
m_k &= \int \ra_d^k\,\text{softmax}(\rva)_\rc\,\N(\rva; \boldsymbol{\mu},\text{diag}(\boldsymbol{\sigma}^2)\,d\rva \\
&= \int_{\ra_d} \ra_d^k\,\N(\ra_d;\mu_d,\sigma_d^2)\int_{\rva \setminus \ra_d} \text{softmax}(\rva)_i \prod_{j\neq i}\N(\ra_j;\mu_j,\sigma_j^2)\,d(\rva\setminus \ra_d) d\ra_d.
\end{align*}
We can reduce the inner integral to the probit approximation by regarding the point distribution $\delta_{\ra_d}$ as the limit of a Gaussian with vanishing variance:
\begin{align*}
    &\int_{\rva \setminus \ra_d} \text{softmax}(\rva)_\rc \prod_{j\neq d}\N(\ra_j;\mu_j,\sigma_j^2)\,d(\rva\setminus \ra_d) \\
= &\int_{\mathbf{a} \setminus \ra_d} \int_{\tilde{a}_d} \delta(\tilde{a}_d - a_d) \, \text{softmax}(a_1, \dots, a_{d-1}, \tilde{a}_d) 
\prod_{j \neq d} \mathcal{N}(a_j; \mu_j, \sigma_j^2) \, d\tilde{a}_d \, d(\rva\setminus \ra_d) \\
= &\int_{\tilde{\mathbf{a}} \setminus \tilde{\ra}_d} \lim_{\sigma \to 0} \int_{\tilde{a}_i} 
\text{softmax}(\tilde{\rva})_\rc\, 
\mathcal{N}(\tilde{a}_d; a_d, \sigma^2) 
\prod_{j \neq d} \mathcal{N}(\tilde{a}_j; \mu_j, \sigma_j^2) \, d\tilde{a}_i \, d\tilde{\mathbf{a}}_i
\end{align*}
By Lebesgue's dominated convergence theorem we obtain equality to
\begin{align*}
    &\lim_{\sigma \to 0} \int_{\tilde{\mathbf{\rva}}} 
\text{softmax}(\tilde{\rva})_\rc \mathcal{N}(\tilde{a}_d; a_d, \sigma^2)
\prod_{j \neq i} \mathcal{N}(\tilde{a}_j; \mu_j, \sigma_j^2) \,d\tilde{\rva}\\
\approx &\lim_{\sigma \to 0} \, \text{softmax}(\vt)_i 
= \text{softmax}(t_1, \dots, t_{d-1}, \ra_d) \quad\text{where}\quad t_j = \begin{cases}
    \mu_j / (1 + \frac{\pi}{8} \sigma_j^2) &\text{for }j\neq d \\ 
    \ra_d / (1 + \frac{\pi}{8} \sigma^2) &\text{for }j=d.
\end{cases}
\end{align*}
Hence, we can approximate $m_k$ by one-dimensional numerical integration of
$$m_k \approx \int_{\ra_d} \ra_d^k\,\N(\ra_d; \mu_d,\sigma_d^2)\,\text{softmax}(t_1,\dots,t_{d-1},\ra_d)\,d\ra_d.$$

\section{Experimental Setup}
\label{appendix:experimental_setup}

\paragraph{Synthetic Data - Depth Scaling:} We generated a dataset of 200 points by randomly sampling $x$ values from the range $[0, 2]$. The true data-generating function was
$$f(x) = 0.5x + 0.2\sin(2\pi \cdot x) + 0.3\sin(4\pi \cdot x).$$
The corresponding $y$ values were sampled by adding Gaussian noise: $f(x) + \mathcal{N}(0, 0.05^2)$. For the architecture, we used a three-layer neural network with the structure:
$$[\text{Linear}(1, 16), \text{LeakyReLU}(0.1), \text{Linear}(16, 16), \text{LeakyReLU}(0.1), \text{Linear}(16, 1)].$$
A four-layer network has one additional $[\text{Linear}(16, 16), \text{LeakyReLU}(0.1)]$ block in the middle, and a five-layer network has two additional blocks. For the regression noise hyperparameter, we used the true noise $\beta^2 = 0.05^2$. The models were trained for 500 iterations over one batch (as all data was processed in a single active batch).

\paragraph{Synthetic Data - Uncertainty Evaluation:} The same data-generation process was used as in the depth-scaling experiment, but this time, $x$ values were drawn from the range $[-0.5, 0.5]$. The network architecture remained the same as the three-layer network, but the width of the layers was increased to 32. We trained 100 networks with different random seeds on the same dataset. We define a $p$-credible interval for $0 \leq p \leq 1$ as:
$$[\text{cdf}^{-1}(0.5 - \frac{p}{2}), \text{cdf}^{-1}(0.5 + \frac{p}{2})].$$
For each credible interval mass $p$ (ranging from 0 to 1 in steps of 0.01), we measured how many of the $p$-credible intervals (across the 100 posterior approximations) covered the true data-generating function. This evaluation was done at each possible $x$ value (ranging from -20 to 20 in steps of 0.05), generating a coverage rate for each combination of $p$ and $x$. For each $p$, we then computed the median for $x > 10$ and the median for $x < -10$. If we correlate the $p$ values with the medians, we found that for the median obtained from positive $x$ values the correlation was 0.96, for negative $x$ it was 0.99, and for the combined set of medians it was $0.9$.

\paragraph{CIFAR-10:} For our CIFAR-10 experiments, we used the default train-test split and trained the following feed-forward network:
\begin{lstlisting}[language=Python]
class Net(nn.Module):
    def __init__(self):
        super(Net, self).__init__()
        self.model = nn.Sequential(
            # Block 1
            nn.Conv2d(3, 32, 3, padding=0),
            nn.LeakyReLU(0.1),
            nn.Conv2d(32, 32, 3, padding=0),
            nn.LeakyReLU(0.1),
            nn.MaxPool2d(2),
            # Block 2
            nn.Conv2d(32, 64, 3, padding=0),
            nn.LeakyReLU(0.1),
            nn.Conv2d(64, 64, 3, padding=0),
            nn.LeakyReLU(0.1),
            nn.MaxPool2d(2),
            # Head
            nn.Flatten(),
            nn.Linear(64 * 5 * 5, 512),
            nn.LeakyReLU(0.1),
            nn.Linear(512, 10),
        )

    def forward(self, x):
        return self.model(x)
\end{lstlisting}
In the case of AdamW and IVON we trained with a cross-entropy loss on the softargmax of the network output.
For our message passing method we used our argmax factor as a training signal instead of softargmax, see \Cref{appendix:message_equation_table}.
The reason is that for softargmax we only have message approximations relying on rather expensive numerical integration.
In our library this factor graph can be constructed via
\begin{lstlisting}[language=Python]
    fg = create_factor_graph([
                size(d.X_train)[1:end-1], # (3, 32, 32)
                # First Block
                (:Conv, 32, 3, 0), # (32, 30, 30)
                (:LeakyReLU, 0.1),
                (:Conv, 32, 3, 0), # (32, 28, 28)
                (:LeakyReLU, 0.1),
                (:MaxPool, 2), # (32, 14, 14)
                # Second Block
                (:Conv, 64, 3, 0), # (64, 12, 12)
                (:LeakyReLU, 0.1),
                (:Conv, 64, 3, 0), # (64, 10, 10)
                (:LeakyReLU, 0.1),
                (:MaxPool, 2), # (64, 5, 5)
                # Head
                (:Flatten,), # (64*5*5 = 1600)
                (:Linear, 512), # (512)
                (:LeakyReLU, 0.1),
                (:Linear, 10), # (10)
                (:Argmax, true)
            ], batch_size)
\end{lstlisting}

For all methods we used a batch size of 128 and trained for 25 epochs with a cosine annealing learning rate schedule.
Concerning hyperparameters: For AdamW we found the standard parameters of $\text{lr}=10^{-3}, \beta_1 = 0.9, \beta_2 = 0.999, \epsilon=10^{-8}\text{ and }\delta = 10^{-4}$ to work best.
For IVON we followed the practical guidelines given in the Appendix of \cite{ivon_shen2024variationallearningeffectivelarge}.

To measure calibration, we used 20 bins that were split to minimize within-bin variance. 
For OOD recognition, we predicted the class of the test examples in CIFAR-10 (in-distribution) and SVHN (OOD) and computed the entropy over softmax probabilities for each example.
We then sort them by negative entropy and test the true positive and false positive rates for each possible (binary) decision threshold. The area under this ROC curve is computed in the same way as for relative calibration.

% \paragraph{Performance Measurements:} We ran four episodes with a total of six iterations (1, 1, 2, 2) for our method, and six episodes for Torch. The training was performed on 3,200 examples, while the inference was conducted on 200,000 examples (by concatenating 20 copies of the original test set). Before running the actual experiments, we performed a warm-up trial, and the final measurements were averaged over three runs. We ran experiments on a g4dn.xlarge AWS instance with 4 vCPUs, 16GB of memory, and an NVIDIA T4 GPU.

\section{Prior Analysis}
\label{appendix:prior_analysis}
The strength of the prior determines the amount of data needed to obtain a useful posterior that fits the data. Our goal is to draw prior means and set prior variances so that the computed variances of all messages are on the order of $\mathcal O(1)$ regardless of network width and depth. It is not entirely clear if this would be a desirable property; after all, adding more layers also makes the network more expressive and more easily able to model functions with very high or low values. However, if we let the predictive prior grow unrestricted, it will grow exponentially, leading to numerical issues. In the following, we analyze the predictive prior under simplifying assumptions to derive a prior initialization that avoids exponential variance explosion. While we fail to achieve this goal, our current prior variances are still informed by this analysis.

In the following, we assume that the network inputs are random variables. Then, the parameters of messages also become random variables, as they are derived from the inputs according to the message equations. Our goal is to keep the expected value of the variance parameter of the outgoing message at a constant size. We also assume that the means of the prior are sampled according to spectral initialization, as described in \Cref{sec:priors}.

\subsubsection*{FirstGaussianLinearLayer - Input is a Constant}
Each linear layer transforms some $d_1$-dimensional input $\mathbf{x}$ to some $d_2$-dimensional output $\mathbf{y}$ according to $\mathbf{y} = W\mathbf{x} + \mathbf{b}$. In the first layer, $\mathbf{x}$ is the input data. For this analysis, we assume each element $x_i$ to be drawn independently from $x_i \sim \mathcal N(0,1)$. Let $\mathbf{x}$ be a $d_1$-dimensional input vector, $\mathbf{m}_w$ be the prior messages from one column of $W$, and $z = \mathbf{w}' \mathbf{x}$ be the vector product before adding the bias.

During initialization of the weight prior, we draw the prior means using spectral parametrization and set the prior variances to a constant:
    $$m_{w_i} = \mathcal N(\mu_{w_i}, \sigma^2_w) ~\text{ with }~ \mu_{w_i} \sim \mathcal N(0,l^2),$$
    $$ l = \frac{1}{\sqrt{k}} \cdot \min(1, \sqrt{\frac{d_2}{d_1}}).$$

By applying the message equations, we then approximate the forward message to the output with a normal distribution
$$m_z = \mathcal N(\mu_z, \sigma^2_z).$$

Because $\sigma^2_z$ depends on the random variables $x_i$, it is also a random variable that follows a scaled chi-squared distribution
\begin{align*}
    \sigma^2_z &= \sum_{i=1}^{d_1} x_i^2 \cdot \sigma^2_w \\
    \sigma^2_z &\sim \chi^2_{d_1} \cdot \sigma^2_w
\end{align*}

and its expected value is
$$\mathbb E[\sigma^2_z] = d_1 \cdot \sigma^2_w.$$

We conclude that we can control the magnitude of the variance parameter by choosing $\mathbb E[\sigma^2_z]$ and setting $\sigma^2_w = \frac{\mathbb E[\sigma^2_z]}{d_1}$.

\subsubsection*{GaussianLinearLayer - Input is a Variable}

In subsequent linear layers, the input $\mathbf{x}$ is not observed and we receive an approximate forward message that consists of independent normal distributions
$$m_{x_i} = \mathcal N(\mu_{x_i}, \sigma^2_{x_i}). $$

Following the message equations, the outgoing forward message to $z$ then has a variance
\begin{align*}
    \sigma^2_z &= \sum_{i=1}^{d_1} (\sigma^2_{x_i} + \mu_{x_i}^2) \cdot (\sigma^2_w + \mu_{w_i}^2) - (\mu_{x_i}^2 * \mu_{w_i}^2) \\
    &= \sum_{i=1}^{d_1} 
        \underbrace{\sigma^2_{x_i} \cdot \sigma^2_w}_{\text{I}}
        + \underbrace{\sigma^2_{x_i} \cdot \mu_{w_i}^2}_{\text{II}}
        + \underbrace{\mu_{x_i}^2 \cdot \sigma^2_w}_{\text{III}}
\end{align*}

The layer's prior variance $\sigma^2_w$ is a constant, whereas all other elements are random variables according to our assumptions. To make further analysis tractable, we also have to assume that the variances $\sigma^2_{x_i}$ of the incoming forward messages are identical constants for all $i$, not random variables. We furthermore assume that the means are drawn i.i.d. from:
\begin{align*}
    \mu_{w_i} &\sim \mathcal N(0, l^2) \\
    \mu_{x_i} &\sim \mathcal N(\mu_{\mu_x}, \sigma^2_{\mu_x}).
\end{align*}

The random variable $\sigma_z^2$ then follows a generalized chi-squared distribution
$$\sigma^2_z \sim \bigg (\sum_{i=1}^{d_1}
    \underbrace{\sigma^2_x \cdot l^2 \cdot \chi^2(1, 0^2)}_{\text{II}}
    + \underbrace{\sigma^2_w \cdot \sigma^2_{\mu_x} \cdot \chi^2(1, \mu^2_{\mu_x})}_{\text{III}}
\bigg )
    + \underbrace{d_1 \cdot \sigma^2_w \cdot \sigma^2_x}_{\text{I}}
$$

and its expected value is
\begin{align*}
    \mathbb E[\sigma^2_z] &= \bigg ( \sum_{i=1}^{d_1} \sigma^2_x \cdot l^2 \cdot (1 + 0^2) + \sigma^2_w \cdot \sigma^2_{\mu_x} \cdot (1 + \mu^2_{\mu_x} ) \bigg ) + d_1 \cdot \sigma^2_w \cdot \sigma^2_x \\
    &= d_1 \cdot \bigg (\sigma^2_x \cdot l^2 + \sigma^2_w \cdot \sigma^2_{\mu_x} \cdot (1 + \mu^2_{\mu_x} ) + \sigma^2_w \cdot \sigma^2_x \bigg )\\
    &= 
        \underbrace{d_1 \cdot \sigma^2_x \cdot l^2}_{\text{II}}
        + \underbrace{d_1 \cdot (\sigma^2_{\mu_x} \cdot (1 + \mu^2_{\mu_x} ) + \sigma^2_x ) \cdot \sigma^2_w}_{\text{I} + \text{III}}.
\end{align*}

As $\sigma^2_w$ has to be positive, we conclude that if we choose $\mathbb E[\sigma^2_z] > d_1 \cdot \sigma^2_x \cdot l^2$, then we can set
$$\sigma^2_w = \frac{\mathbb E[\sigma^2_z] - d_1 \cdot \sigma^2_x \cdot l^2}{d_1 \cdot (\sigma^2_{\mu_x} \cdot (1 + \mu^2_{\mu_x} ) + \sigma^2_x )}.$$

We know (or choose) $d_1$, $l^2$, and $\mathbb E[\sigma^2_z]$, but we require values for $\sigma^2_x$, $\mu^2_{\mu_x}$, and $\sigma^2_{\mu_x}$ to be able to choose $\sigma^2_w$. We will find empirical values for these parameters in the next section.

\subsubsection*{Empirical parameters + LeakyReLU}
% - We have a formula for the first layer available
% - We assume that our MLP will follow the architecture linear, leaky,  linear,  leaky,  ..., linear
% - Now we want to find estimates for the three parameters that are needed to set the prior variances of the inner linear layers
% - Specifically, we assume that we have pre-activation messages with means that are iid drawn from \N(0, 1) and with variances that are t (which is the target variance of the subsequent layer).
% - As LeakyReLU's message equations are complex, we use experimental data instead of analysis for finding approximations to the three parameters
% - We find that relationship between the LeakyReLU's leak and $\mu_{\mu_x}$ (specifically, the average over means of the output messages) is approximately linear, while the relationship between the leak and $\sigma^2_{\mu_x}$ (the empirical variance of the output messages) and the relationship between the leak and $\mu_{\sigma^2_x}$ (the mean over variances of the output messages) are both approximately quadratic.
% -> Using simulation, we find coefficients for these relationships with an error of circa $5 \cdot 10^{-5}$.
% - We can then use the coefficients for identifying the three parameters needed to inform the prior variance of the inner linear layers.
% - We opt to keep the target variance of each layer at $t=2$ and use a leak of 0.1, which then leads to setting $\sigma^2_x = 0.8040586726631379$ and $\sigma^2_{\mu_x} \cdot (1 + \mu^2_{\mu_x} ) = 0.44958619556324186$.
% -> We can now set the priors of our entire MLP!

To inform the choice of the prior variances of the inner linear layers, we also need to analyze LeakyReLU. We assume the network is an MLP that alternates between linear layers and LeakyReLU. As the message equations of LeakyReLU are too complicated for analysis, we instead use empirical approximation. Let $m_a = \N(\mu_a, \sigma^2_a)$ be an incoming message (from the pre-activation variable to LeakyReLU). We assume that $\sigma^2_a = t$ is a constant and that $\mu_a \sim \N(0, 1)$ is a random variable. By sampling multiple means and then computing the outgoing messages (after applying LeakyReLU), we can approximate the average variance of the outgoing messages, as well as the average and empirical variance over means of the outgoing messages.

We computed these statistics for 101 different leak settings with 100 million samples each, and found that the relationship between leak and $\mu_{\mu_x}$ (average mean of the outgoing message) is approximately linear, while the relationships between leak and $\sigma^2_{\mu_x}$ or $\mu_{\sigma^2_x}$ are approximately quadratic. Using these samples, we fitted coefficients with an error margin below $5 \cdot 10^{-5}$. For our network, we chose a target variance of $1.5$ and a leak of $0.1$, resulting in
\begin{align*}
    \sigma^2_x &= 0.8040586726631379\\
    \sigma^2_{\mu_x} \cdot (1 + \mu^2_{\mu_x} ) &= 0.44958619556324186.
\end{align*}

These values are sufficient for now setting the prior variances of the inner linear layer according to the equations above. Finally, we set the prior variance of the biases to $0.5$, so that the output of each linear layer achieves an overall target prior predictive variance of approximately $t = 1.5 + 0.5 = 2.0$.

\subsubsection*{Results in practice}
In practice, we found that the variance of the predictive posterior still goes up exponentially with the depth of the network despite our derived prior choices. However, if we lower the prior variance further to avoid this explosion, the network is overly restricted and unable to obtain a good fit during training. We therefore set the prior variances as outlined here, but acknowledge that choosing a good prior is still an unsolved problem.

\section{Tables of Message Equations}
\label{appendix:message_equation_table}

In the following, we provide tables summarizing all message equations used throughout our model. The tables are divided into three categories: linear algebra operations (\Cref{tab:message_equations}), training signals (\Cref{tab:message_equations_training_signals}), and activation functions (\Cref{tab:message_equations3}). Each table contains the relevant forward and backward message equations, along with illustrations of the corresponding factor graph where necessary. These summaries serve as a reference for the mathematical operations performed during inference and training, and they will be valuable for factor graph modeling across various domains beyond neural networks.

\begin{table}[htbp]
    \centering
    \begin{NiceTabular}{|c|l c|}
        \toprule
        \multirow{4.5}{*}{\rotatebox{90}{Product}} & \multicolumn{2}{c}{$\begin{gathered}
            \begin{tikzpicture}
	\begin{pgfonlayer}{nodelayer}
        \node [style={variable_node}, minimum size=7mm] (0) at (-5.75, 0.875) {$\ra$};
        \node [style={variable_node}, minimum size=7mm] (1) at (-5.75, -0.875) {$\rb$};
        
		\node [style=factor] (2) at (0.0, 0.0) {};
        \node [anchor=south] at (0.0, 0.2) {$\delta(\rz - \ra\rb)$};
        \node [style={variable_node}, minimum size=7mm] (3) at (5.75, 0.0) {$\rz$};

        % \node [style=none] (4) at (-10.75, 0.875) {};
        % \node [style=none] (5) at (-8.75, 1.375) {$\N(\hat \mu_\ra, \hat \sigma_\ra^2)$};
        % \node [style=none] (6) at (-10.75, -0.875) {};
        % \node [style=none] (7) at (-8.75, -0.375) {$\N(\hat \mu_\rb, \hat \sigma_\rb^2)$};
        
        \node [style=none] (8) at (10.75, 0.0) {};
        \node [style=none] (9) at (8.75, 0.5) {$\N(\mu_\rz, \sigma_\rz^2)$};
	\end{pgfonlayer}
	\begin{pgfonlayer}{edgelayer}
		\draw (0) to (2);
		\draw (1) to (2);
            \draw[thick, deoxygenated_blue, arrows = {-Triangle}] (2) to (1);
            \draw[thick, oxygenated_red, arrows = {-Triangle}] (2) to (3);
        % \draw [style={incoming_message}] (4.center) to (0);
        % \draw [style={incoming_message}] (6.center) to (1);
        \draw [style={incoming_message}] (8.center) to (3);
	\end{pgfonlayer}
\end{tikzpicture}
        \end{gathered}$} \\
        \cmidrule(lr){2-3}
        & \tikz[]{\draw[thick, oxygenated_red, arrows = {-Triangle}] (0.0,0.0) to (1.5,0.0);} & $\begin{aligned}
            \textcolor{oxygenated_red}{\mu} &= \E[\ra] \E[\rb] &
            \textcolor{oxygenated_red}{\sigma^2} &= \E[\ra^2] \E[\rb^2] - \E[\ra]^2\E[\rb]^2
        \end{aligned}$\\
        \cmidrule(lr){2-3}
        & \tikz[]{\draw[thick, deoxygenated_blue, arrows = {-Triangle}] (1.5,0.0) to (0.0,0.0);} & $\begin{aligned}
            \textcolor{deoxygenated_blue}{\tau} &= \tau_z  \E[\ra] & \textcolor{deoxygenated_blue}{\rho} &= \rho_z \E[\ra^2]
        \end{aligned}$ \\
        \midrule

        \multirow{9.5}{*}{\rotatebox{90}{Weighted Sum}} & \multicolumn{2}{c}{$\begin{gathered}
            \begin{tikzpicture}
	\begin{pgfonlayer}{nodelayer}
        \node [style={known_variable}] (0) at (-5.75, 3.25) {$a_1$};
        \node [style={variable_node}, minimum size=7mm] (1) at (-5.75, 2.0) {$\rb_1$};

        \node [style={known_variable}] (2) at (-5.75, -1) {$a_d$};
	\node [style={variable_node}, minimum size=7mm] (3) at (-5.75, -2.25) {$\rb_d$};
        
	\node [style=none] (6) at (-5.75, 0.7) {$\vdots$};
	\node [style=factor] (7) at (1.0, 0.5) {};
        \node [anchor=south] at (0.95, 1.3) {$\delta\left(\rz - \va^\intercal \rvb \right)$};
        \node [style={variable_node}, minimum size=7mm] (8) at (7.75, 0.5) {$z$};

        \node [style=none] (9) at (-10.75, 2.0) {};
        \node [style=none] (10) at (-8.75, 2.75) {$\N(\mu_1, \sigma_1^2)$};
        \node [style=none] (11) at (-10.75, -2.25) {};
        \node [style=none] (12) at (-8.75, -1.5) {$\N(\mu_d, \sigma_d^2)$};
        \node [style=none] (13) at (12.75, 0.5) {};
        \node [style=none] (14) at (10.75, 1.25) {$\N(\mu_z, \sigma_z^2)$};
	\end{pgfonlayer}
	\begin{pgfonlayer}{edgelayer}
		\draw (0) to (7);
		\draw (1) to (7);
		\draw (2) to (7);
		\draw[thick, deoxygenated_blue, arrows = {-Triangle}] (7) to (3);
		\draw[thick, oxygenated_red, arrows = {-Triangle}] (7) to (8);
        \draw [arrows = {-Triangle}] (9.center) to (1);
        \draw [arrows = {-Triangle}] (11.center) to (3);
        \draw [arrows = {-Triangle}] (13.center) to (8);
	\end{pgfonlayer}
\end{tikzpicture}
        \end{gathered}$} \\
        \cmidrule(lr){2-3}
        & \tikz[]{\draw[thick, oxygenated_red, arrows = {-Triangle}] (0.0,0.0) to (1.5,0.0);} & $\begin{aligned}
            \textcolor{oxygenated_red}{\mu} &= \va^\intercal \vmu & \textcolor{oxygenated_red}{\sigma^2} &= \sum_{i=1}^d a_i^2 \sigma^2_i
        \end{aligned}$\\
        \cmidrule(lr){2-3}
        & \tikz[]{\draw[thick, deoxygenated_blue, arrows = {-Triangle}] (1.5,0.0) to (0.0,0.0);} & $\begin{aligned}
            \textcolor{deoxygenated_blue}{\mu} &= (\mu_z - \textcolor{oxygenated_red}{\mu} + a_d\mu_d)/ a_d &
            \textcolor{deoxygenated_blue}{\tau} &= a_d\frac{ \tau_z - \rho_z(\textcolor{oxygenated_red}{\mu} - a_d\mu_d)}{1 +  \rho_z ( \textcolor{oxygenated_red}{\sigma^2} - a_d^2\sigma^2_d)} \\
            \textcolor{deoxygenated_blue}{\sigma^2} &= (\sigma_z^2 + \textcolor{oxygenated_red}{\sigma^2} - a_d^2\sigma_d^2) / a_d^2 & \textcolor{deoxygenated_blue}{\rho} &= \frac{a_d^2  \rho_z}{1 +  \rho_z ( \textcolor{oxygenated_red}{\sigma^2} - a_d^2\sigma^2_d)}
        \end{aligned}$
        \\
        \midrule

        \multirow{10.5}{*}{\rotatebox{90}{Inner Product}} & \multicolumn{2}{c}{$\begin{gathered}
            \begin{tikzpicture}
	\begin{pgfonlayer}{nodelayer}
        \node [style={variable_node}] (0) at (-5.75, 3.25) {$\ra_1$};
        \node [style={variable_node}, minimum size=7mm] (1) at (-5.75, 2.0) {$\rb_1$};
        \node [style={variable_node}] (16) at (1.75, 2.625) {$\rp_1$};

        \node [style={variable_node}] (2) at (-5.75, -1) {$\ra_d$};
	\node [style={variable_node}, minimum size=7mm] (3) at (-5.75, -2.25) {$\rb_d$};
        \node [style={variable_node}] (17) at (1.75, -1.625) {$\rp_d$};
        
	\node [style=none] (6) at (-5.75, 0.7) {$\vdots$};
	\node [style=factor] (7) at (-2.0, 2.65) {};
        \node [anchor=south] at (-2.0, 1.0) {$\delta\left(\rp_1 - \ra_1\rb_1 \right)$};
	\node [style=factor] (15) at (-2.0, -1.65) {};
        \node [anchor=south] at (-2.0, -1.2) {$\delta\left(\rp_d - \ra_d\rb_d \right)$};
        \node [style={variable_node}, minimum size=7mm] (8) at (9.25, 0.5) {$\rz$};
        \node [style=factor] (18) at (5.5, 0.5) {};
        \node [anchor=south] at (6.5, -2.0) {$\delta\left(\rz - \sum_{i=1}^d \rp_i\right)$};

        % \node [style=none] (9) at (-10.75, 2.0) {};
        % \node [style=none] (10) at (-8.75, 2.75) {$\N(\mu_1, \sigma_1^2)$};
        % \node [style=none] (11) at (-10.75, -2.25) {};
        % \node [style=none] (12) at (-8.75, -1.5) {$\N(\mu_d, \sigma_d^2)$};
        \node [style=none] (13) at (14.5, 0.5) {};
        \node [style=none] (14) at (12.3, 1.25) {$\N(\mu_\rz, \sigma_\rz^2)$};
	\end{pgfonlayer}
	\begin{pgfonlayer}{edgelayer}
		\draw (0) to (7);
		\draw (1) to (7);
		\draw (2) to (15);
            \draw (7) to (16);
            \draw (15) to (17);
            \draw (16) to (18);
            \draw (17) to (18);
		\draw[thick, deoxygenated_blue, arrows = {-Triangle}] (15) to (3);
		\draw[thick, oxygenated_red, arrows = {-Triangle}] (18) to (8);
        % \draw [arrows = {-Triangle}] (9.center) to (1);
        % \draw [arrows = {-Triangle}] (11.center) to (3);
        \draw [arrows = {-Triangle}] (13.center) to (8);
	\end{pgfonlayer}
\end{tikzpicture}
        \end{gathered}$} \\
        \cmidrule(lr){2-3}
        & \tikz[]{\draw[thick, oxygenated_red, arrows = {-Triangle}] (0.0,0.0) to (1.5,0.0);} & $\begin{aligned}
            \textcolor{oxygenated_red}{\mu} &= \E[\ra]^\intercal \E[\rb] &
            \textcolor{oxygenated_red}{\sigma^2} &= \sum_{i=1}^d \E[\ra_i^2]\E[\rb_i^2] - \E[\ra_i]^2\E[\rb_i]^2
        \end{aligned}$\\
        \cmidrule(lr){2-3}
        & \tikz[]{\draw[thick, deoxygenated_blue, arrows = {-Triangle}] (1.5,0.0) to (0.0,0.0);} & $ \begin{aligned}
            \textcolor{deoxygenated_blue}{\tau_{b_i}} &= \frac{ \tau_z - \rho_z (\textcolor{oxygenated_red}{\mu} -  \E[\ra_d]\E[\rb_d])}{\rho^*_i} \E[\ra_d]\\
            \textcolor{deoxygenated_blue}{\rho_{\rb_i}} &=  \frac{\rho_z}{\rho^*_i}\E[\ra_d^2] \\
            \text{where}~~~ \rho^*_i &= 1 + \rho_z ( \textcolor{oxygenated_red}{\sigma^2} - \E[\ra_d^2]\E[\rb_d^2] + \E[\ra_d]^2\E[\rb_d]^2) 
        \end{aligned}$ \\
        \bottomrule
    \end{NiceTabular}
    \caption{Message equations for linear algebra: Calculating backward messages in natural parameters is preferable as it handles edge cases like $a_d = 0$ or $\rho_z = 0$ where location-scale equations are ill-defined. This approach also enhances numerical stability by avoiding division by very small quantities. Note that the inner product messages are simply compositions of the product and weighted sum messages with $\ra_i = 1,~i=1,\dots,d.$}
    \label{tab:message_equations}
\end{table}

\begin{table}[htbp]
    \centering
    \begin{NiceTabular}{|c|l c|}
        \toprule
        \multirow{5}{*}{\rotatebox{90}{Regression}} & \multicolumn{2}{c}{$\begin{gathered}
            \begin{tikzpicture}
	\begin{pgfonlayer}{nodelayer}
        \node [style={variable_node}, minimum size=7mm] (0) at (-3.75, 0.0) {$\ra$};
		\node [style=factor] (1) at (1.0, 0.0) {};
        \node [anchor=north] at (1.0, -0.2) {$\N(\ra; y, \beta^2)$};
        \node [style={variable_node}, minimum size=7mm] (2) at (5.75, 0.0) {$\ry$};
        \node [style={known_variable}, minimum size=6mm] (3) at (1.0, 3.0) {\scalebox{0.9}{$\beta$}};

        \node [style=none] (4) at (-8.75, 0.0) {};
        \node [style=none] (5) at (-6.75, 0.5) {$\N(\mu_{\ra}, \sigma_{\ra}^2)$};
	\end{pgfonlayer}
	\begin{pgfonlayer}{edgelayer}
		\draw[thick, deoxygenated_blue, arrows = {-Triangle}] (1) to (0);
		\draw[thick, oxygenated_red, arrows = {-Triangle}] (1) to (2);
        \draw (1) to (3);
        \draw [arrows = {-Triangle}] (4.center) to (0);
	\end{pgfonlayer}
\end{tikzpicture}
        \end{gathered}$} \\
        \cmidrule(lr){2-3}
        & \tikz[]{\draw[thick, oxygenated_red, arrows = {-Triangle}] (0.0,0.0) to (1.5,0.0);} & $\begin{aligned}
            \textcolor{oxygenated_red}{\mu} &= \mu_\ra & \textcolor{oxygenated_red}{\sigma^2} &= \sigma^2_\ra + \beta^2
        \end{aligned}$\\
        \cmidrule(lr){2-3}
        &  \tikz[]{\draw[thick, deoxygenated_blue, arrows = {-Triangle}] (1.5,0.0) to (0.0,0.0);} & $\begin{gathered}
            \msg{f}{\ra}(\ra) =\begin{cases}
                \N(\ra; y, \beta^2) &\text{if }\ry\text{ is known (training branch)}\\
                1 &\text{if }\ry\text{ is unknown (prediction branch)}
            \end{cases}  
        \end{gathered}$ \\
        \midrule

        \multirow{13.5}{*}{\rotatebox{90}{Softmax}} & \multicolumn{2}{c}{$\begin{gathered}
            \begin{tikzpicture}
	\begin{pgfonlayer}{nodelayer}
        \node [style={variable_node}, minimum size=7mm] (0) at (-6.75, 2.625) {$\ra_1$};
        \node [style=none] (2) at (-6.75, 0.75) {$\vdots$};
        \node [style={variable_node}, minimum size=7mm] (3) at (-6.75, -1.775) {$\ra_d$};
        
        \node [style=factor] (4) at (0.0, 0.6) {};
        \node [] at (0.0,2.0) {$\text{softmax}(\rva)_\rc$};

        \node [style={variable_node}, minimum size=7mm] (5) at (6.75, 0.6) {$\rc$};
        \node [style=none] (16) at (6.75, -1) {$\rc \in \{1,\dots,d\}$};

        \node [style=none] (9) at (-11.75, 2.625) {};
        \node [style=none] (10) at (-9.75, 3.125) {$\N(\mu_1, \sigma_1^2)$};
        
        \node [style=none] (11) at (-11.75, 0.875) {};
        
        \node [style=none] (13) at (-11.75, -1.775) {};
        \node [style=none] (14) at (-9.75, -1.275) {$\N(\mu_d, \sigma_d^2)$};

	\end{pgfonlayer}
	\begin{pgfonlayer}{edgelayer}
		\draw (0) to (4);
            \draw (3) to (4);
            \draw[thick, deoxygenated_blue, arrows = {-Triangle}] (4) to (3);
            \draw[thick, oxygenated_red, arrows = {-Triangle}] (4) to (5);
        \draw [arrows = {-Triangle}] (9.center) to (0);
        \draw [arrows = {-Triangle}] (13.center) to (3);
        % \draw [style={incoming_message}] (15.center) to (5);
        % \draw [style={incoming_message}] (17.center) to (6);
        % \draw [style={incoming_message}] (19.center) to (8);
	\end{pgfonlayer}
\end{tikzpicture}
        \end{gathered}$} \\
        \cmidrule(lr){2-3}
        & \tikz[]{\draw[thick, oxygenated_red, arrows = {-Triangle}] (0.0,0.0) to (1.5,0.0);} & $\begin{aligned}
            \msg{f}{\rc}(i) &= \int \text{softmax}(\rva)_i \N(\rva; \vmu, \text{diag}(\boldsymbol{\sigma})^2)\,d\rva \\
            &\approx \text{softmax}(\vt)_i~~~(\text{\cite{daxberger2022laplacereduxeffortless}})\\
            \text{where}~~~t_j &= \frac{\mu_j}{1 + \frac{\pi}{8}\sigma_j^2}~~~\text{and}~~~j=1,\dots,d
        \end{aligned}$ 
        \\
        \cmidrule(lr){2-3}
        & \tikz[]{\draw[thick, deoxygenated_blue, arrows = {-Triangle}] (1.5,0.0) to (0.0,0.0);} & $ \begin{aligned}
            
            \msg{f}{\ra_d}(\ra_d) &= \frac{\N(\ra_d; m_1 / m_0, m_2 / m_0 - ( m_1/m_0)^2)} {\msg{\ra_d}{f}(\ra_d)} \\
            \text{where}~~~m_k &= \int_{\ra_d} \ra_d^k\,\N(\ra_d; \mu_d, \sigma^2_d)\,\text{softmax}(t_1, ..., t_{d-1}, \ra_d)_\rc\,d\ra_d \\
            &\text{is approximated via numerical integration}
        \end{aligned}$ \\
        \midrule

        \multirow{17.5}{*}{\rotatebox{90}{Argmax}} & \multicolumn{2}{c}{$\begin{gathered}
            \begin{tikzpicture}
	\begin{pgfonlayer}{nodelayer}
        \node [style={variable_node}, minimum size=7mm] (0) at (-6.75, 2.625) {$\ra_1$};
        \node [style=none] (2) at (-6.75, 1.0) {$\vdots$};
        \node [style={variable_node}, minimum size=7mm] (3) at (-6.75, -1.0) {$\ra_d$};
        \node [style={variable_node}, minimum size=7mm] (16) at (1.25, 0.78) {$\rc$};
        \node [style={variable_node}, minimum size=7mm] (17) at (5.25, 2.625) {$\rz_1$};
        \node [style={variable_node}, minimum size=7mm] (18) at (5.25, -1.0) {$\rz_d$};
        
        \node [style=factor] (4) at (-2.75, 2.625) {};
        \node [anchor=south] at (-2.5, 2.8) {$\delta(\rz_1 - (\ra_\rc - \ra_1))$};
        \node [style=none] (5) at (-2.75, 1.0) {$\vdots$};
        \node [style=factor] (6) at (-2.75, -1.0) {};
        \node [anchor=north] at (-2.5, -1.2) {$\delta(\rz_d - (\ra_\rc - \ra_d))$};
        \node [style=factor] (19) at (9.25, 2.625) {};
        \node [anchor=north] at (9.25, 4.0) {$\1_{\rz_1\,\geq\,0}$};
        \node [style=factor] (20) at (9.25, -1.0) {};
        \node [anchor=north] at (9.25, -1.2) {$\1_{\rz_d\,\geq\,0}$};

        \node [style=none] (9) at (-11.75, 2.625) {};
        \node [style=none] (10) at (-9.75, 3.125) {$\N(\mu_1, \sigma_1^2)$};
        
        \node [style=none] (13) at (-11.75, -1.0) {};
        \node [style=none] (14) at (-9.75, -0.5) {$\N(\mu_d, \sigma_d^2)$};
	\end{pgfonlayer}
	\begin{pgfonlayer}{edgelayer}
		\draw (0) to (4);
		\draw (3) to (4);
        \draw (0) to (6);
        \draw (3) to (6);
        \draw (4) to (16);
        \draw (4) to (17);
        \draw (6) to (18);
        \draw (17) to (19);
        \draw (18) to (20);
        \draw[thick, oxygenated_red, arrows = {-Triangle}] (6) to (16);

        \draw [arrows = {-Triangle}] (9.center) to (0);
        \draw [arrows = {-Triangle}] (13.center) to (3);
	\end{pgfonlayer}
\end{tikzpicture}
            \hspace{7mm}
        \end{gathered}$} \\
        \cmidrule(lr){2-3}
        & \tikz[]{\draw[thick, oxygenated_red, arrows = {-Triangle}] (0.0,0.0) to (1.5,0.0);} & $\begin{aligned}
            &\int \1_{\rz_d\,\geq 0}\delta(\rz_d - (\ra_\rc - \ra_d))\N(\ra_\rc; \mu_\rc, \sigma_\rc^2)\N(\ra_d;\mu_d,\sigma_d^2)\,d\ra_\rc d\ra_d d\rz\\
            = &\begin{cases}
                1 &\text{ for }\rc = d\\
                \Pr[\ra_\rc \geq \ra_d] = \phi(0; \mu_d - \mu_\rc, \sigma_d^2 + \sigma_\rc^2) &\text{ for }\rc \neq d
            \end{cases}
        \end{aligned}$\\
        \cmidrule(lr){2-3}
        & \tikz[]{\draw[thick, deoxygenated_blue, arrows = {-Triangle}] (1.5,0.0) to (0.0,0.0);} & \parbox[]{0.9\textwidth}{If $\rc$ is known, many edges become constant and can be omitted. Assume w.l.o.g. $\rc = d$, then $\ra_d$ is connected to $d-1$ factors and all other $\ra_i$ to only one each. The messages to $\ra_1,\dots,\ra_d$ follow from the weighted sum factor, given Gaussian approximations of the messages from $\rz_i$. We derive these by moment-matching the marginals of $\rz_i$ (see Building Block \ref{appendix:building_block_relu_moments}) and dividing by the message from the weighted sum factor. To stabilize training, we regularize the variance of $\msg{f}{\ra_i}$ by a factor of $\phi(0; \mu_\rc - \mu_i, \sigma^2_{i} + \sigma^2_\rc)$ and multiply $\msg{f}{\ra_i}(\ra_i) \text{ by } \N(\ra_\rc; 1 \text{ if }i=\rc\text{ else }-1, \gamma^2)$, effectively mixing in one-hot regression factors during training.} \\

        \bottomrule
    \end{NiceTabular}
    \caption{Message equations for training signals. Note that the backward messages only apply in the case in which the target is known, i.e., on the training branches. On the prediction branch we only do foward passes.}
    \label{tab:message_equations_training_signals}
\end{table}

\begin{table}[htbp]
    \centering
    \begin{NiceTabular}{|c|l c|}
        \toprule
        \multirow{10.5}{*}{\rotatebox{90}{Auxiliary Equations}}
        & \multicolumn{2}{c}{ $\begin{aligned}
            \text{ReLUMoment}_k(\mu, \sigma^2) &= \begin{cases}
                \E[\text{ReLU}(\ra)] \text{ with } \ra \sim \N(\mu, \sigma^2)                  &~~~~~~\text{for } k=1\\
                \E[\text{ReLU}^2(\ra)] \text{ with } \ra \sim \N(\mu, \sigma^2)                 &~~~~~~\text{for } k=2
            \end{cases}\\
            &= \begin{cases}
                \sigma \varphi(x) + \mu \phi(x)                     &~~~~~~~~~\text{for } k=1\\
                \sigma\mu \varphi(x) + (\sigma^2 + \mu^2)\phi(x)    &~~~~~~~~~\text{for } k=2
            \end{cases} \\
            &\text{where }\varphi \text{ and }\phi \text{ denote the pdf and cdf of $\N(0,1)$, respectively.}
        \end{aligned}$} \\
        \cmidrule(lr){2-3}
        & \multicolumn{2}{c}{ $\begin{gathered}
            \begin{aligned}
                \zeta_k(\mu_1,\sigma_1, \mu_2, \sigma_2) &:= \int_0^\infty \ra^k\N(\ra;\mu_1,\sigma^2_1)\N(\ra; \mu_2, \sigma^2_2)\,d\ra \\ 
                &= \N(\mu_1; \mu_2, \sigma^2_1 + \sigma^2_2) \cdot  \begin{cases}
                    \text{ReLUMoment}_k(\mu_m, \sigma^2_m)              &~~~~~~~~~~~~~~~\text{for } k=1,2\\
                    \phi(\mu_m / \sigma_m)                         &~~~~~~~~~~~~~~~\text{for } k=0
                \end{cases}
            \end{aligned} \\
            \text{with }
            \tau_m = \frac{\mu_1}{\sigma^2_1} + \frac{\mu_2}{\sigma^2_2}, ~~~
            \rho_m = \frac{1}{\sigma^2_1} + \frac{1}{\sigma^2_2}, ~~~
            \mu_m = \frac{\tau_m}{\rho_m}, ~~~\text{and}~
            \sigma^2 = \frac{1}{\rho_m} \\
            \text{See Building Block \ref{building_block:product_of_gaussians_positive} for the derivation of this equation.}
        \end{gathered}$} \\
        \midrule

        \multirow{8}{*}{\rotatebox{90}{LeakyReLU}} & \multicolumn{2}{c}{$\begin{gathered}
            \begin{tikzpicture}
	\begin{pgfonlayer}{nodelayer}
        \node [style={variable_node}, minimum size=7mm] (0) at (-5.75, 0.0) {$\ra$};
		\node [style=factor] (1) at (1.0, 0.0) {};
        \node [anchor=south] at (1.0, 0.2) {$\delta(\ra - \text{LeakyReLU}_{\alpha}(\rz))$};
        \node [style={variable_node}, minimum size=7mm] (2) at (7.75, 0.0) {$\rz$};

        \node [style=none] (3) at (-10.75, 0.0) {};
        \node [style=none] (4) at (-8.75, 0.5) {$\N(\mu_{\ra}, \sigma_{\ra}^2)$};
        
        \node [style=none] (5) at (12.75, 0.0) {};
        \node [style=none] (6) at (10.75, 0.5) {$\N(\mu_\rz, \sigma_\rz^2)$};
	\end{pgfonlayer}
	\begin{pgfonlayer}{edgelayer}
        \draw[thick, deoxygenated_blue, arrows = {-Triangle}] (1) to (0);
        \draw[thick, oxygenated_red, arrows = {-Triangle}] (1) to (2);
        \draw [arrows = {-Triangle}] (3.center) to (0);
        \draw [arrows = {-Triangle}] (5.center) to (2);
	\end{pgfonlayer}
\end{tikzpicture}
        \end{gathered}$} \\
        \cmidrule(lr){2-3}
        & \multicolumn{2}{c}{ $\begin{aligned}
            &\text{We use marginal approximation while:}\\
            &~~\text{1. The outputs are finite and not NaN} \\
            &~~\text{2. Forward message: Precision of } \msg{f}{\rz} \text{ is } \geq \text{ precision of } \msg{\ra}{f} \text{, and }  m_0 > 10^{-8} \\
            &~~\text{3. Backward message: It has worked well to require } (\tau_\rz > 0) \lor ( \rho_\rz > 2 \cdot 10^{-8}) \\
            &\text{Otherwise, we fall back to direct message approximation (forward) or } \sG(0,0) \text{ (backward).}
        \end{aligned}$} \\
        \midrule

        \multirow{-1.65}{*}{\rotatebox{90}{Direct}}
        & \multicolumn{2}{c}{ $\begin{aligned}
            & \\[-3mm]
             \textcolor{oxygenated_red}{\mu} &= (1 - \alpha) \cdot \text{ReLUMoment}_1( \mu_\ra,  \sigma^2_\ra) + \alpha\cdot  \mu_\ra \\
             \textcolor{oxygenated_red}{\sigma^2} &= (1 - \alpha^2) \cdot \text{ReLUMoment}_2 ( \mu_\ra,  \sigma^2_\ra) + \alpha^2 \cdot ( \sigma^2_\ra +  \mu_\ra^2) -  \textcolor{oxygenated_red}{\mu}^2. \\[-3mm]
            &~
        \end{aligned}$ }\\
        \midrule

        \multirow{-2.4}{*}{\rotatebox{90}{Marginal}}
        & \multicolumn{2}{c}{ $\begin{gathered}
            \msg{f}{\rz}(\rz) = \frac{\N(\rz; \frac{ m_1}{ m_0}, \frac{ m_2}{ m_0} - ( \frac{ m_1}{ m_0}) ^2)} {\msg{\rz}{f}(\rz)}\\
            \begin{aligned}
                
                 \text{where}\quad m_k = & ~ (-1)^k \cdot \zeta_k (-  \mu_a,  \sigma^2_a, ~~ -\alpha \cdot  \mu_\rz, \alpha^2 \cdot  \sigma^2_\rz) + \zeta_k( \mu_a,  \sigma^2_a, ~~  \mu_\rz,  \sigma^2_\rz)
            \end{aligned} \\[3mm]
            \text{To compute the marginal backward message, set } \alpha_{\text{back}} = \alpha^{-1} \\
            \text{ and swap } \msg{a}{f} \text{ and } \msg{\rz}{f} \text{ in the equation}
        \end{gathered}$ }\\
        
        \bottomrule
    \end{NiceTabular}
    \caption{Message equations for LeakyReLU with ReLU as the special case $\alpha=0$.}
    \label{tab:message_equations3}
\end{table}

\end{document}